\documentclass[journal]{IEEEtran}
\usepackage{hyperref}
\usepackage{url}
\usepackage{graphicx}
\usepackage{booktabs} 
\usepackage{adjustbox}
\usepackage{amsmath}
\usepackage{colortbl} 
\usepackage{xcolor} 
\usepackage[skins]{tcolorbox}
\usepackage[utf8]{inputenc}
\usepackage{dirtytalk}

\usepackage{amsmath}
\usepackage{mathtools}
\usepackage{amsfonts}
\DeclareMathOperator*{\argmax}{arg\,max}
\usepackage{cite}
\begin{document}
\title{Natural Language Processing Advancements By Deep Learning: A Survey}

\author{Amirsina~Torfi,~\IEEEmembership{Member,~IEEE,}
        Rouzbeh A.~Shirvani,
        ~Yaser Keneshloo,
        Nader Tavaf,
        and~Edward~A.~Fox,~\IEEEmembership{Fellow,~IEEE}
\thanks{Amirsina Torfi, Yaser Keneshloo, and Edward A. Fox were with the Department
of Computer Science, Virginia Polytechnic Institute and State University, Blacksburg,
VA, 24060 USA e-mail: (amirsina.torfi@gmail.com, yaserkl@vt.edu, fox@vt.edu). Rouzbeh A. Shirvani is an independent researcher, e-mail: (rouzbeh.asghari@gmail.com). Nader Tavaf was with the University of Minnesota Twin Cities, Minneapolis, MN, 55455 USA e-mail: (tavaf001@umn.edu).}
}

\markboth{Torfi\MakeLowercase{ et. al.}, NLP Advancements by Deep Learning}%
{}

\maketitle

\begin{abstract}
Natural Language Processing (NLP) helps empower intelligent machines by enhancing a better understanding of the human language for linguistic-based human-computer communication. Recent developments in computational power and the advent of large amounts of linguistic data have heightened the need and demand for automating semantic analysis using data-driven approaches. The utilization of data-driven strategies is pervasive now due to the significant improvements demonstrated through the usage of deep learning methods in areas such as Computer Vision, Automatic Speech Recognition, and in particular, NLP. This survey categorizes and addresses the different aspects and applications of NLP that have benefited from deep learning. It covers core NLP tasks and applications, and describes how deep learning methods and models advance these areas.~We further analyze and compare different approaches and state-of-the-art models.
\end{abstract}

\IEEEpeerreviewmaketitle

\section{Introduction}
\IEEEPARstart{N}{atural} Language Processing (NLP) is a sub-discipline of computer science providing a bridge between natural languages and computers.
It helps empower machines to understand, process, and analyze human language~\cite{manning1999foundations}. 
NLP's significance as a tool aiding comprehension of human-generated data is a logical consequence of the context-dependency of data.
Data becomes more meaningful through a deeper understanding of its context, which in turn facilitates text analysis and mining.
NLP enables this with the communication structures and patterns of humans.

Development of NLP methods is increasingly reliant on data-driven approaches which help with building more powerful and robust models~\cite{zhang2015character,cho2014learning, Wu2020}. 
Recent advances in computational power, as well as greater availability of big data, enable deep learning, one of the most appealing approaches in the NLP domain~\cite{collobert2008unified,zhang2015character,cho2014learning},~especially given that deep learning has already demonstrated superior performance in adjoining fields like Computer Vision~\cite{karpathy2014large,oquab2014learning,shrivastava2017learning, Voulodimos2018, Arai2020} and Speech Recognition~\cite{graves2014towards,amodei2016deep, kamath2019deep}. 
These developments led to a paradigm shift from traditional to novel data-driven approaches aimed at advancing NLP.
The reason behind this shift was simple: new approaches are more promising regarding results, and are easier to engineer.

As a sequitur to remarkable progress achieved in adjacent disciplines utilizing deep learning methods, deep neural networks have been applied to various NLP tasks,~including part-of-speech tagging~\cite{santos2014learning,plank2016multilingual,manning2011part, deshmukh2020deep},~named entity recognition~\cite{lample2016neural,chiu2015named,lample2016neural, yadav2019survey, li2020survey}, and semantic role labeling~\cite{zhou2015end,marcheggiani2017simple,he2017deep, he2019syntax}.
Most of the research efforts in deep learning associated with NLP applications involve either \textit{supervised learning\footnote{Learning from training data to predict the type of new unseen test examples by mapping them to known pre-defined labels.}} or \textit{unsupervised learning\footnote{Making sense of data without sticking to specific tasks and supervisory signals.}}.



This survey covers the emerging role of deep learning in the area of NLP, across a broad range of categories.
The research presented in~\cite{young2018recent} is primarily focused on architectures, with little discussion of applications. More recent works~\cite{Kang2020, Wu2020} are specific to certain applications or certain sub-fields of NLP~\cite{li2020survey}.
Here we build on previous works by describing the challenges, opportunities, and evaluations of the impact of applying deep learning to NLP problems. 

This survey has six sections, including this introduction.
\textbf{Section 2} lays out the theoretical dimensions of NLP and artificial intelligence, and looks at deep learning as an approach to solving real-world problems.
It motivates this study by addressing the question: Why use deep learning in NLP?
The \textbf{third section} discusses fundamental concepts necessary to understand NLP, covering exemplary issues in representation, frameworks, and machine learning. 
The \textbf{fourth section} summarizes benchmark datasets employed in the NLP domain.
 \textbf{Section 5} focuses on some of the NLP applications where deep learning has demonstrated significant benefit.
 Finally, \textbf{Section 6} provides a conclusion, also addressing some open problems and promising areas for improvement.
\section{Background}

NLP has long been viewed as one aspect of artificial intelligence (AI), since understanding and generating natural language are high-level indications of intelligence.
Deep learning is an effective AI tool, so we next situate deep learning in the AI world.
After that we explain motivations for applying deep learning to NLP.

\subsection{Artificial Intelligence and Deep Learning}

There have been ``islands of success'' where big data are processed via AI capabilities to produce information to achieve critical operational goals (e.g., fraud detection).
Accordingly, scientists and consumers anticipate enhancement across a variety of applications.
However, achieving this requires understanding of AI and its mechanisms and means (e.g., algorithms).
Ted Greenwald, explaining AI to those who are not AI experts, comments: \textit{"Generally AI is anything a computer can do that formerly was considered a job for a human"} \cite{breenwald2018wsj}. 

An AI goal is to extend the capabilities of information technology (IT) from those to (1) generate, communicate, and store data, to also (2) process data into the knowledge that decision makers and others need~\cite{sivarajah2017critical}.
One reason is that the available data volume is increasing so rapidly that it is now impossible for people to process all available data.
This leaves two choices:~(1) much or even most existing data must be ignored or (2) AI must be developed to process the vast volumes of available data into the essential pieces of information that decision-makers and others can comprehend. Deep learning is a bridge between the massive amounts of data and AI.

\subsubsection{Definitions}
Deep learning refers to \textit{applying deep neural networks to massive amounts of data to learn a procedure aimed at handling a task}.
The task can range from simple classification to complex reasoning. 
In other words, deep learning is \textit{a set of mechanisms} ideally capable of deriving an optimum solution to any problem given a sufficiently extensive and relevant input dataset.
Loosely speaking, deep learning is detecting and analyzing important structures/features in the data aimed at formulating a solution to a given problem. 
Here, AI and deep learning meet.
One version of the goal or ambition behind AI is  enabling a machine \textit{to outperform what the human brain does}.
Deep learning is a means to this end. 

\subsubsection{Deep Learning Architectures}
Numerous deep learning architectures have been developed in different research areas, e.g., in NLP applications employing recurrent neural networks (RNNs)~\cite{lipton2015critical}, convolutional neural networks (CNNs)~\cite{kim2014convolutional}, and more recently, recursive neural networks~\cite{socher2011parsing}.
We focus our discussion on a review of the essential models, explained in relevant seminal publications.

\begin{figure}[ht]
\centering
    \includegraphics[scale=0.13]{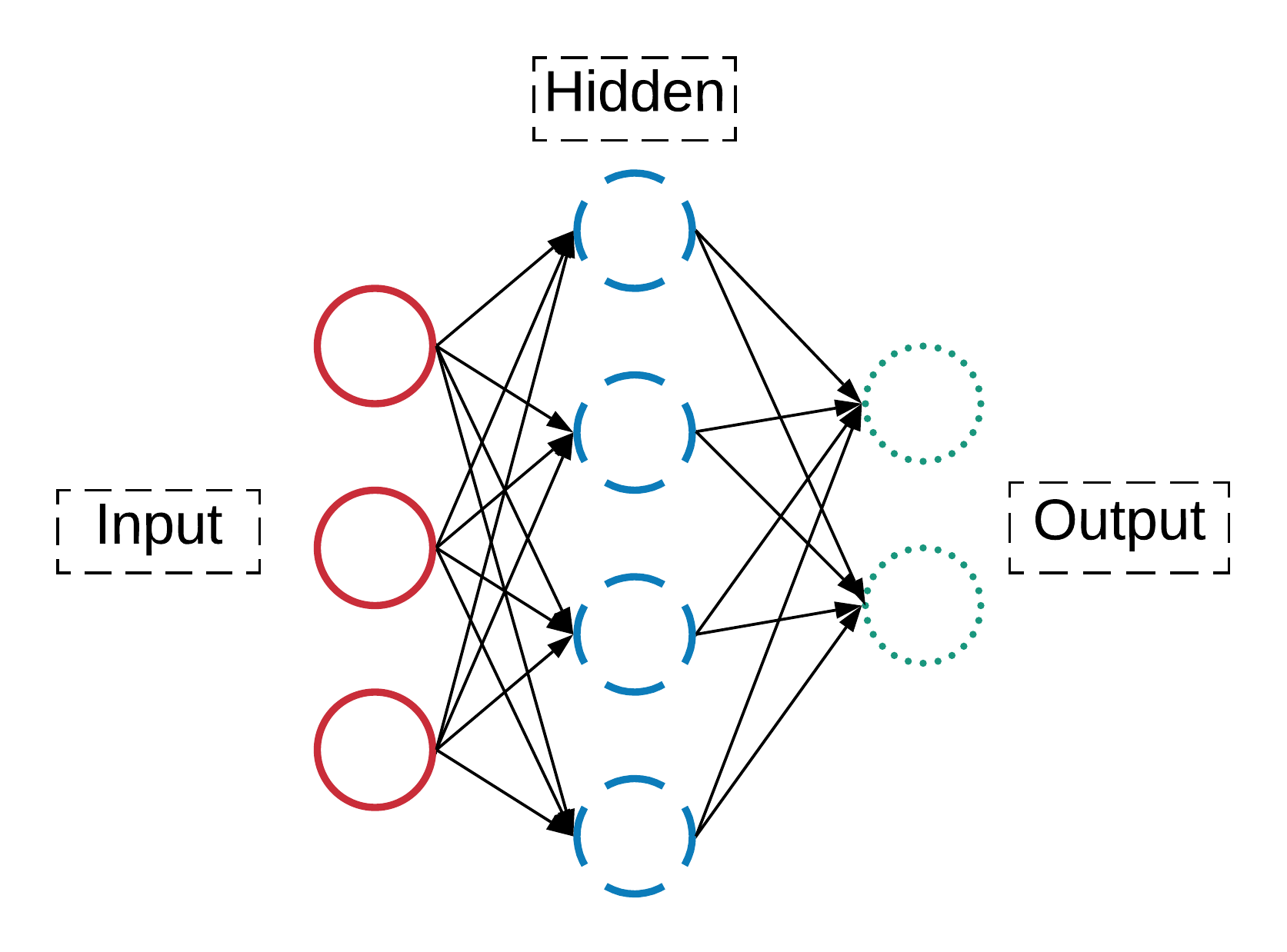}
    \caption{The general architecture of a MLP.} \label{fig:mlp}
\end{figure}

\textbf{Multi Layer Perceptron:}
A \textit{multilayer perceptron} (MLP) has at least three layers~(input, hidden, and output layers).
A layer is simply a collection of neurons operating to transform information from the previous layer to the next layer.
In the MLP architecture, the neurons in a layer do not communicate with each other.
An MLP employs nonlinear activation functions.
Every node in a layer connects to all nodes in the next layer,~creating a fully connected network~(Fig.~\ref{fig:mlp}).
MLPs are the simplest type of \textit{Feed-Forward Neural Networks}~(FNNs).
FNNs represent a general category of neural networks in which the connections between the nodes do not create any cycle, i.e., in a FNN there is no cycle of information flow.

\textbf{Convolutional Neural Networks:} Convolutional neural networks~(CNNs), whose architecture is inspired by the human visual cortex, are a subclass of feed-forward neural networks. 
CNNs are named after the underlying mathematical operation, \textit{convolution}, which yields a measure of the interoperability of its input functions.~Convolutional neural networks are usually employed in situations where data is or needs to be represented with a 2D or 3D data map.~In the data map representation, the proximity of data points usually corresponds to their information correlation. 

In convolutional neural networks where the input is an image,  the data map indicates that image pixels are highly correlated to their neighboring pixels.~Consequently, the convolutional layers have 3 dimensions: width, height, and depth.~That assumption possibly explains why the majority of research efforts dedicated to CNNs are conducted in the Computer Vision field~\cite{krizhevsky2012imagenet}.

\begin{figure*}[ht]
  \centering
  \includegraphics[width=0.95\textwidth]{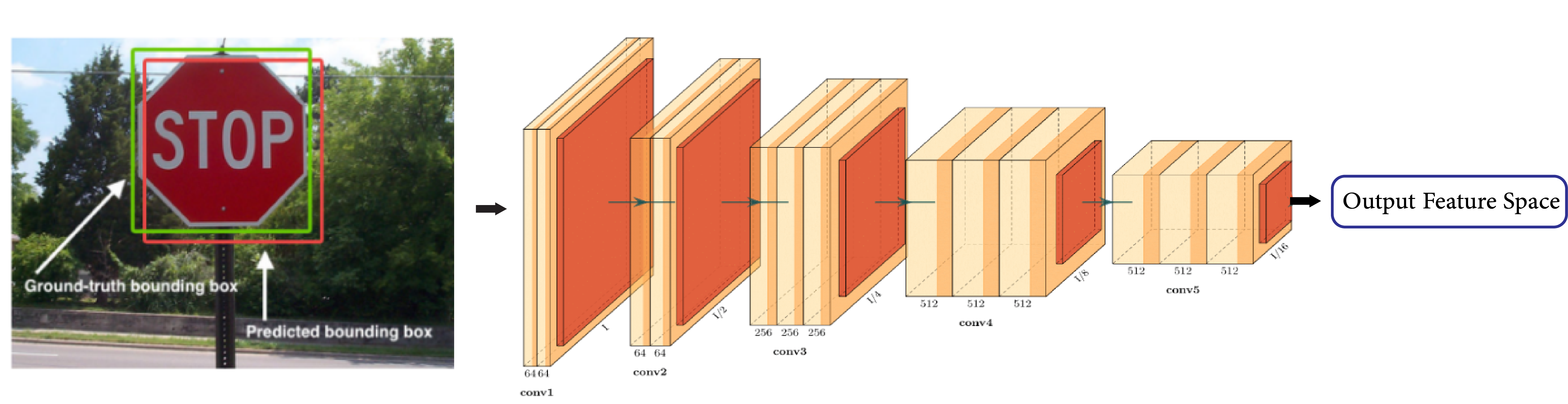}
  \caption{A typical CNN architecture for object detection. The network provides a feature representation with attention to the specific region of an image (example shown on the left) that contains the object of interest. Out of the multiple regions represented (see an ordering of the image blocks, giving image pixel intensity, on the right) by the network, the one with the highest score will be selected as the main candidate.}\label{fig:typical-cnn}
\end{figure*}

A CNN takes an image represented as an array of numeric values.
After performing specific mathematical operations, it represents the image in a new output space.
This operation is also called feature extraction, and helps to capture and represent key image content.
The extracted features can be used for further analysis, for different tasks.
One example is image classification, which aims to categorize images according to some predefined classes.
Other examples include determining which objects are present in an image and where they are located.
See Fig.~\ref{fig:typical-cnn}.

In the case of utilizing CNNs for NLP,~the inputs are sentences or documents represented as matrices.~Each row of the matrix is associated with a language element such as a word or a character.~The majority of CNN architectures learn word or sentence representations in their training phase.~A variety of CNN architectures were used in various classification tasks such as Sentiment Analysis and Topic Categorization \cite{kim2014convolutional,dos2014deep,johnson2014effective,johnson2015semi}.
CNNs were employed for Relation Extraction and Relation Classification as well~\cite{zeng2014relation,nguyen2015relation}.

\begin{figure}[ht]
\centering
    \includegraphics[totalheight=2.6cm]{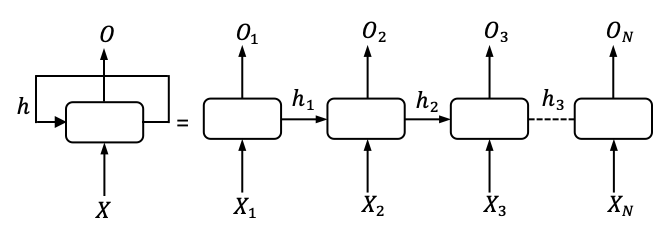}
    \caption{Recurrent Neural Network (RNN), summarized on the left, expanded on the right, for $N$ timesteps, with $X$ indicating input, $h$ hidden layer, and $O$ output} \label{fig:rnn}
\end{figure}

\textbf{Recurrent Neural Network:} If we line up a sequence of FNNs and feed the output of each FNN as an input to the next one, a recurrent neural network (RNN) will be constructed.
Like FNNs, layers in an RNN can be categorized into input, hidden, and output layers.
In discrete time frames, sequences of input vectors are fed as the input, one vector at a time, e.g., after inputting each batch of vectors, conducting some operations and updating the network weights, the next input batch will be fed to the network. 
Thus, as shown in Fig.~\ref{fig:rnn}, at each time step we make predictions and use parameters of the current hidden layer as input to the next time step.

Hidden layers in recurrent neural networks can carry information from the past, in other words, memory.
This characteristic makes them specifically useful for applications that deal with a sequence of inputs such as language modeling~\cite{mikolov2010recurrent}, i.e., representing language in a way that the machine understands.
This concept will be described later in detail.

RNNs can carry rich information from the past.
Consider the sentence: ``Michael Jackson was a singer; some people consider him King of Pop.''
It's easy for a human to identify \textit{him} as referring to Michael Jackson. 
The pronoun \textit{him} happens seven words after \textit{Michael Jackson;} capturing this dependency is one of the benefits of RNNs, where the hidden layers in an RNN act as memory units.
Long Short Term Memory Network (LSTM)~\cite{hochreiter1997long} is one of the most widely used classes of RNNs.
LSTMs try to capture even long time dependencies between inputs from different time steps.
Modern Machine Translation and Speech Recognition often rely on LSTMs.

\begin{figure}[ht]
\centering
    \includegraphics[totalheight=3.2cm]{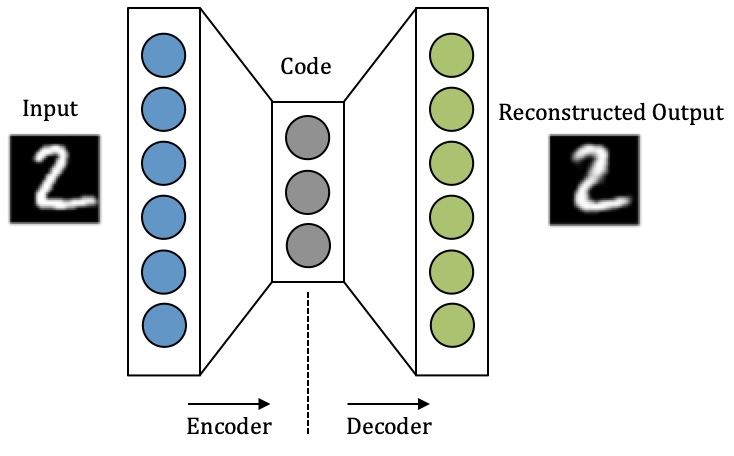}
    \caption{Schematic of an Autoencoder} \label{fig:autoencode}
\end{figure}

\textbf{Autoencoders:} \textit{Autoencoders} implement unsupervised methods in deep learning.
They are widely used in dimensionality reduction\footnote{Dimensionality reduction is an unsupervised learning approach which is the process of reducing the number of variables that were used to represent the data by identifying the most crucial information.} or NLP applications which consist of sequence to sequence modeling~(see Section~\ref{sec:seq2seq}~\cite{mikolov2010recurrent}.
Fig.~\ref{fig:autoencode} illustrates the schematic of an Autoencoder.
Since autoencoders are unsupervised, there is no label corresponding to each input.
They aim to learn a code representation for each input.
The encoder is like a feed-forward neural network in which the input gets encoded into a vector (code).
The decoder operates similarly to the encoder, but in reverse, i.e., constructing an output based on the encoded input.
In data compression applications, we want the created output to be as close as possible to the original input.
Autoencoders are \textit{lossy}, meaning the output is an approximate reconstruction of the input.

\begin{figure}[ht]
\centering
    \includegraphics[totalheight=6cm]{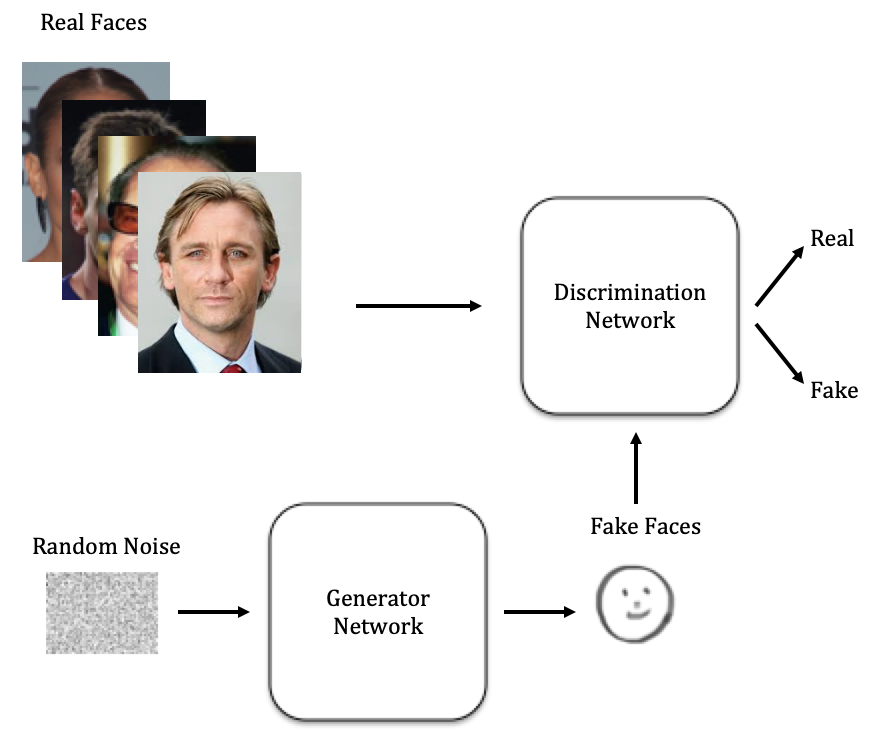}
    \caption{Generative Adversarial Networks} \label{fig:gan}
\end{figure}

\textbf{Generative Adversarial Networks:} Goodfellow~\cite{goodfellow2014generative} introduced \textit{Generative Adversarial Networks~(GANs)}.
As shown in Fig.~\ref{fig:gan}, a GAN is a combination of two neural networks, a discriminator and a generator.
The whole network is trained in an iterative process.
First, the generator network generates a fake sample.
Then the discriminator network tries to determine whether this sample (ex.: an input image) is real or fake, i.e., whether it came from the real training data (data used for building the model) or not.
The goal of the generator is to fool the discriminator in a way that the discriminator believes the artificial (i.e., generated) samples synthesized by the generator are real.

This iterative process continues until the generator produces samples that are indistinguishable by the discriminator.
In other words, the probability of classifying a sample as fake or real becomes like flipping a fair coin for the discriminator.
The goal of the generative model is to capture the distribution of real data while the discriminator tries to identify the fake data.
One of the interesting features of GANs~(regarding being generative) is: once the training phase is finished, there is no need for the discrimination network, so we solely can work with the generation network.
In other words, having access to the trained generative model is sufficient.

Different forms of GANs has been introduced, e.g., Sim GAN~\cite{shrivastava2017learning}, Wasserstein GAN~\cite{arjovsky2017wasserstein}, info GAN~\cite{chen2016infogan}, and DC GAN~\cite{radford2015unsupervised}.
In one of the most elegant GAN implementations \cite{karras2017progressive}, entirely artificial, yet almost perfect, celebrity faces are generated; the pictures are not real, but fake photos produced by the network. GAN's have since received significant attention in various applications and have generated astonishing result~\cite{Tavaf2021}.
In the NLP domain, GANs often are used for text generation~\cite{yu2017seqgan,li2017adversarial}.

\subsection{Motivation for Deep Learning in NLP}

Deep learning applications are predicated on the choices of (1) feature representation and (2) deep learning algorithm alongside  architecture.
These are associated with data representation and learning structure, respectively.
For data representation, surprisingly, there usually is a disjunction between what information is thought to be important for the task at hand, versus what representation actually yields good results.
For instance, in sentiment analysis, lexicon semantics, syntactic structure, and context are assumed by some linguists to be of primary significance.
Nevertheless, previous studies based on the bag-of-words~(BoW) model demonstrated acceptable performance~\cite{pang2002thumbs}. The bag-of-words model~\cite{harris1954distributional}, often viewed as the vector space model, involves a representation which accounts only for the words and their frequency of occurrence.
BoW ignores the order and interaction of words, and treats each word as a unique feature.
BoW disregards syntactic structure, yet provides decent results for what some would consider syntax-dependent applications.
This observation suggests that simple representations, when coupled with large amounts of data, may work as well or better than more complex representations.~These findings corroborate the argument in favor of the importance of deep learning algorithms and architectures.

Often the progress of NLP is bound to effective language modeling.
A goal of statistical language modeling is the probabilistic representation of word sequences in language, which is a complicated task due to the curse of dimensionality.
The research presented in~\cite{bengio2003neural} was a breakthrough for language modeling with neural networks aimed at overcoming the curse of dimensionality by \textbf{(1)} learning a distributed representation of words and \textbf{(2)} providing a probability function for sequences. 

A key challenge in NLP research, compared to other domains such as Computer Vision, seems to be the complexity of achieving an in-depth representation of language using statistical models.
A primary task in NLP applications is to provide a representation of texts, such as documents.
This involves feature learning, i.e., extracting meaningful information to enable further processing and analysis of the raw data.

Traditional methods begin with time-consuming hand-crafting of features, through careful human analysis of a specific application, and are followed by development of algorithms to extract and utilize instances of those features.
On the other hand, deep supervised feature learning methods are highly data-driven and can be used in more general efforts aimed at providing a robust data representation. 

Due to the vast amounts of unlabeled data, unsupervised feature learning is considered to be a crucial task in NLP.~Unsupervised feature learning is, in essence, learning the features from unlabeled data to provide a low-dimensional representation of a high-dimensional data space.~Several approaches such as K-means clustering and principal component analysis have been proposed and successfully implemented to this end.~With the advent of deep learning and abundance of unlabeled data, unsupervised feature learning becomes a crucial task for representation learning, a precursor in NLP applications.~Currently, most of the NLP tasks rely on annotated data, while a preponderance of unannotated data further motivates research in leveraging deep data-driven unsupervised methods.

Given the potential superiority of deep learning approaches in NLP applications, it seems crucial to perform a comprehensive analysis of various deep learning methods and architectures with particular attention to NLP applications. 

\section{Core Concepts in NLP}
\subsection{Feature Representation}

Distributed representations are a series of compact, low dimensional representations of data, each representing some distinct informative property.~For NLP systems, due to issues related to the atomic representation of the symbols, it is imperative to learn word representations.

At first, let's concentrate on how the features are represented, and then we focus on different approaches for learning word representations.
The encoded input features can be characters, words~\cite{socher2011parsing},~sentences~\cite{le2014distributed}, or other linguistic elements.
Generally, it is more desirable to provide a compact representation of the words than a sparse one.

\begin{figure}[ht]
  \centering
  \includegraphics[height=3cm]{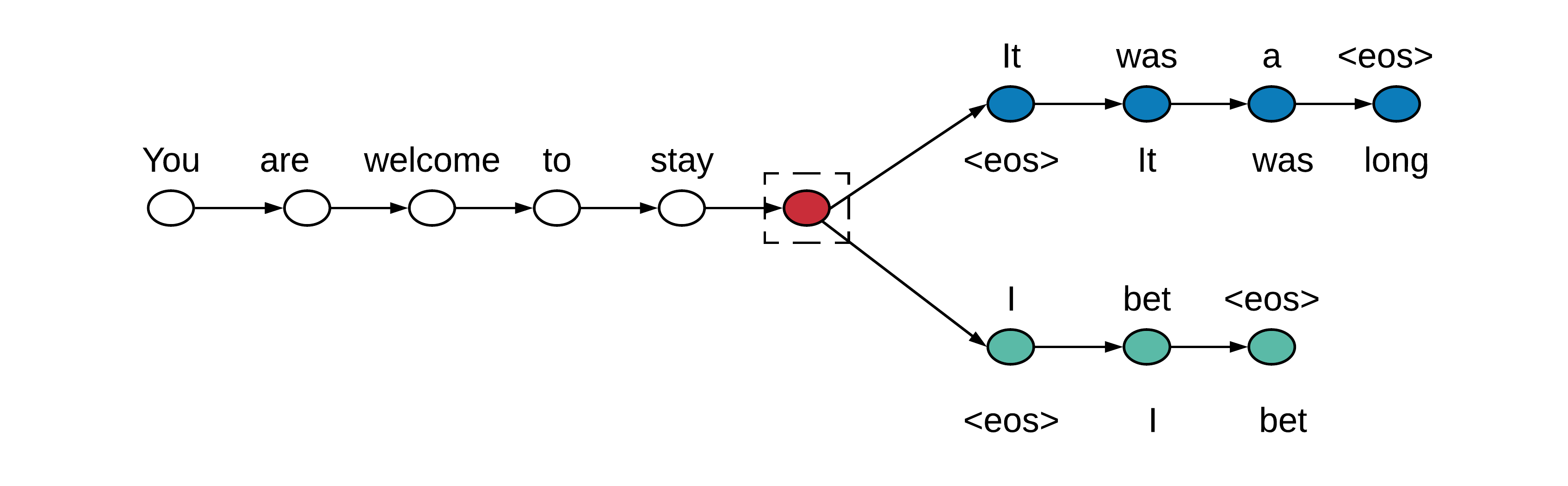}
  \caption{Considering a given sequence, the skip-thought model generates the surrounding sequences using the trained encoder.
  The assumption is that the surrounding sentences are closely related, contextually.}\label{fig:skipthought}
\end{figure}

How to select the structure and level of text representation used to be an unresolved question.
After proposing the word2vec approach~\cite{mikolov2013distributed}, subsequently, doc2vec was proposed in \cite{le2014distributed} as an unsupervised algorithm and was called Paragraph Vector~(PV).
The goal behind PV is to learn fixed-length representations from variable-length text parts such as sentences and documents.
One of the main objectives of doc2vec is to overcome the drawbacks of models such as BoW and to provide promising results for applications such as text classification and sentiment analysis.
A more recent approach is the skip-thought model which applies word2vec at the sentence-level~\cite{kiros2015skip}.
By utilizing an encoder-decoder architecture, this model generates the surrounding sentences using the given sentence~(Fig.~\ref{fig:skipthought}). Next, let's investigate different kinds of feature representation.

\subsubsection{One-Hot Representation}

In one-hot encoding,~each unique element that needs to be represented has its dimension which results in a very high dimensional,~very sparse representation.~Assume the words are represented with the one-hot encoding method.
Regarding representation structure, there is no meaningful connection between different words in the feature space.
For example, highly correlated words such as `ocean' and `water' will not be closer to each other~(in the representation space) compared to less correlated pairs such as `ocean' and `fire.'
Nevertheless, some research efforts present promising results using one-hot encoding~\cite{zhang2015character}.

\subsubsection{Continuous Bag of Words}

Continuous Bag-of-Words model (CBOW) has frequently been used in NLP applications.
CBOW tries to predict a word given its surrounding context, which usually consists of a few nearby words~\cite{mikolov2013efficient}.
CBOW is neither dependent on the sequential order of words nor necessarily on probabilistic characteristics.
So it is not generally used for language modeling.
This model is typically trained to be utilized as a pre-trained model for more sophisticated tasks.
An alternative to CBOW is the weighted CBOW (WCBOW)~\cite{lebanon2015riemannian} in which different vectors get different weights reflective of relative importance in context.
The simplest example can be document categorization where features are words and weights are TF-IDF scores~\cite{leskovec2014mining} of the associated words.

\subsubsection{Word-Level Embedding}
Word embedding is a learned representation for context elements in which, ideally, words with related semantics become highly correlated in the representation space. 
One of the main incentives behind word embedding representations is the high generalization power as opposed to sparse, higher dimensional representations~\cite{goldberg2017neural}.
Unlike the traditional bag-of-words model in which different words have entirely different representations regardless of their usage or collocations, learning a distributed representation takes advantage of word usage in context to provide similar representations for semantically correlated words.
There are different approaches to create word embeddings.
Several research efforts,~including~\cite{mikolov2013efficient,mikolov2013distributed}, used random initialization by uniformly sampling random numbers with the objective of training an efficient representation of the model on a large dataset.
This setup is intuitively acceptable for initialization of the embedding for common features such as part-of-speech tags.
However, this may not be the optimum method for representation of less frequent features such as individual words.
For the latter, pre-trained models, trained in a supervised or unsupervised manner, are usually leveraged for increasing the performance.

\subsubsection{Character-Level Embedding}

The methods mentioned earlier are mostly at higher levels of representation.
Lower-level representations such as character-level representation require special attention as well, due to their simplicity of representation and the potential for correction of unusual character combinations such as misspellings~\cite{zhang2015character}.
For generating character-level embeddings, CNNs have successfully been utilized~\cite{santos2014learning}.

Character-level embeddings have been used in different NLP applications~\cite{wehrmann2017character}.
One of the main advantages is the ability to use small model sizes and represent words with lower-level language elements~\cite{santos2014learning}.
Here word embeddings are models utilizing CNNs over the characters.
Another motivation for employing character-level embeddings is the out-of-vocabulary word~(OOV) issue which is usually encountered when, for the given word, there is no equivalent vector in the word embedding. 
The character-level approach may significantly alleviate this problem.
Nevertheless, this approach suffers from a weak correlation between characters and semantic and syntactic parts of the language.
So, considering the aforementioned pros and cons of utilizing character-level embeddings, several research efforts tried to propose and implement higher-level approaches such as using sub-words~\cite{bojanowski2016enriching} to create word embeddings for OOV instances as well as creating a semantic bridge between the correlated words~\cite{botha2014compositional}.


\subsection{Seq2Seq Framework}\label{sec:seq2seq}
Most underlying frameworks in NLP applications rely on sequence-to-sequence (seq2seq) models in which not only the input but also the output is represented as a sequence.
These models are common in various applications including machine translation\footnote{The input is a sequence of words from one language (e.g., English) and the output is the translation to another language (e.g., French).}, text summarization\footnote{The input is a complete document (sequence of words) and the output is a summary of it (sequence of words).}, speech-to-text, and text-to-speech applications\footnote{The input is an audio recording of a speech (sequence of audible elements) and the output is the speech text (sequence of words).}.

The most common seq2seq framework is comprised of an encoder and a decoder.
The encoder ingests the sequence of input data and generates a mid-level output which is subsequently consumed by the decoder to produce the series of final outputs.
The encoder and decoder are usually implemented via a series of Recurrent Neural Networks or LSTM~\cite{hochreiter1997long} cells.

The encoder takes a sequence of length $T$, $X = \{x_1, x_2, \cdots , x_T\}$, where $x_t \in V = \{1, \cdots , |V|\}$ is the representation of a single input coming from the vocabulary $V$, and then generates the output state $h_t$.
Subsequently, the decoder takes the last state from the encoder, i.e., $h_t$, and starts generating an output of size $L$, $Y^{\prime} = \{y^{\prime}_1, y^{\prime}_2, \cdots , y^{\prime}_L \}$, based on its current state, $s_t$, and the ground-truth output $y_t$. In different applications, the decoder could take advantage of more information such as a context vector~\cite{see2017get} or intra-attention vectors~\cite{paulus2017deep} to generate better outputs.

One of the most widely training approaches for seq2seq models is called \textit{Teacher Forcing}~\cite{bengio2015scheduled}. Let us define $y = \{y_1, y_2, \cdots , y_L \}$ as the ground-truth output sequence correspondent to a given input sequence $X$.
The model training based on the maximum-likelihood criterion employs the following cross-entropy~(CE) loss minimization:

\begin{equation}
    \mathcal{L}_{CE}=-\sum_{t=1}^{L}\log{p_{\theta}(y_{t}|y_{t-1},s_t,X)}
\end{equation}
where $\theta$ is the parameters of the model optimized during the training.

Once the model is optimized using the cross-entropy loss, it can generate an entire sequence as follows.
Let $\hat{y}_t$ denote the output generated by the model at time $t$.
Then, the next output is generated by:
\begin{equation}
\hat{y}_{t} = \argmax_{y} p_{\theta}(y|\hat{y}_{t-1}, s_{t})
\label{eq:inf}
\end{equation}

In NLP applications, one can improve the output by using beam search to find a reasonably good output sequence~\cite{cho2014learning}.
During beam search, rather than using $\verb|argmax|$ for selecting the best output, we choose the top $K$ outputs at each step,  generate $K$ different paths for the output sequence, and finally choose the one that provides better performance as the final output.
Although, there has been some recent studies~\cite{goyal2018continuous,kool2019stochastic} on improving the beam search by incorporating a similar mechanism during training of them model, studying this is outside the scope of this paper.

Given a series of the ground-truth output $Y$ and the generated model output $\hat{Y}$, the model performance is evaluated using a task-specific measures such as $\textrm{ROUGE}$~\cite{lin2004rouge}, $\textrm{BLEU}$~\cite{papineni2002bleu}, and $\textrm{METEOR}$~\cite{banerjee2005meteor}.
As an example, $\textrm{ROUGE}_L$, which is an evaluation metric in NLP tasks, uses the largest common sub-string between ground-truth $Y$ and model output $\hat{Y}$ to evaluate the generated output.

\subsection{Reinforcement Learning in NLP}
Although the seq2seq models explained in Section~\ref{sec:seq2seq} achieve great successes w.r.t. traditional methods, there are some issues with how these models are trained.
Generally speaking, seq2seq models like the ones used in NLP applications face two issues: (1) \textit{exposure bias} and (2) \textit{inconsistency between training time and test time measurements}~\cite{keneshloo2018deep}.

Most of the popular seq2seq models are minimizing cross-entropy loss as their optimization objective via Teacher Forcing~(Section~\ref{sec:seq2seq}). In teacher forcing, during the training of the model, the decoder utilizes two inputs, the former decoder output state $s_{t-1}$ and the ground-truth input $y_t$, to determine its current output state $s_t$. Moreover, it employs them to create the next token, i.e., $\hat{y}_t$. However, at test time, the decoder fully relies on the previously created token from the model distribution. As the ground-truth data is not available, such a step is necessary to predict the next action. Henceforth, in training, the decoder input is coming from the ground truth, while, in the test phase, it relies on the previous prediction. 

This \textit{exposure bias}~\cite{ranzato2015sequence} induces error growth through output creation at the test phase. One approach to remedy this problem is to remove the ground-truth dependency in training by solely relying on model distribution to minimize the cross-entropy loss. Scheduled sampling~\cite{bengio2015scheduled} is one popular method to handle this setback.
During scheduled sampling, we first pre-train the model using cross-entropy loss and then slowly replace the ground-truth with samples the model generates.

The second obstacle with seq2seq models is that, when training is finished using the cross-entropy loss, it is typically evaluated using non-differentiable measures such as $\textrm{ROUGE}$ or $\textrm{METEOR}$. This will form an inconsistency between the training objective and the test evaluation metric. Recently, it has been demonstrated that both of these problems can be tackled by utilizing techniques from reinforcement learning~\cite{keneshloo2018deep}.

Among most of the well-known models in reinforcement learning, policy gradient techniques~\cite{zaremba2015reinforcement} such as the REINFORCE algorithm~\cite{williams1992simple} and actor-critic based models such as value-based iteration~\cite{sutton2018reinforcement}, and Q-learning~\cite{watkins1992q}, are among the most common techniques used in deep learning in NLP.

Using the model predictions~(versus the ground-truth) for the sequence to sequence modeling and generation, at training time, was initially introduced by Daume \textit{et al.}~\cite{daume2009search}. 
According to their approach, SEARN, the structured prediction can be characterized as one of the reinforcement learning cases as follows:
\textit{The model employs its predictions to produce a sequence of actions (words sequences). Then, at each time step, a greedy search algorithm is employed to learn the optimal action, and the policy will be trained to predict that particular action}. 

\begin{figure}[ht]
    \centering
    \includegraphics[width=0.99\columnwidth]{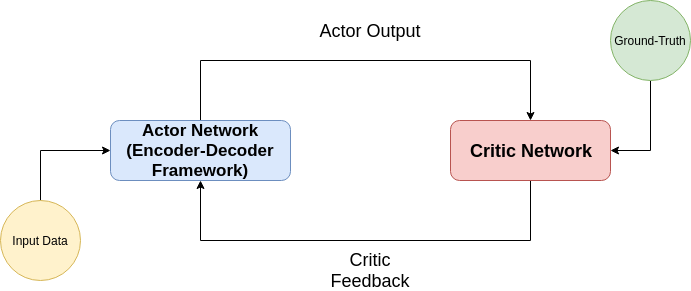}
    \caption{A simple Actor-Critic framework.}
    \label{fig:ac}
\end{figure}

In Actor-Critic training, the actor is usually the same neural network used to generate the output, while the critic is a regression model that estimates how the actor performed on the input data.
The actor later receives the feedback from the critic and improves its actions.
Fig~\ref{fig:ac} shows this framework.
It is worth noting that action in most of the NLP-related applications is like selecting the next output token while the state is the decoder output state at each stage of decoding.
These models have mostly been used for robotic~\cite{levine2016end} and Atari games~\cite{mnih2014recurrent} due to the small action space in these applications.
However, when we use them in NLP applications, they face multiple challenges.
The action space in most of the NLP applications could be defined as the number of tokens in the vocabulary (usually between 50K to 150K tokens).
Comparing this to the action space in a simple Atari game, which on average has less than 20 actions~\cite{mnih2014recurrent}, shows why these Actor-Critic models face difficulties when applied to NLP applications.
A major challenge is the massive action space in NLP applications, which not only causes difficulty for the right action selection, but also will make the training process very slow.
This makes the process of finding the best Actor-Critic model very complicated and model convergence usually requires a lot of tweaks to the models.

\section{Datasets}

Many different researchers for different tasks use benchmark datasets, such as those discussed below.
Benchmarking in machine learning refers to the assessment of methods and algorithms, comparing those regarding their capability to learn specific patterns.
Benchmarking aids validation of a new approach or practice, relative to other existing methods. 

Benchmark datasets typically take one of three forms. 
\begin{enumerate}
    \item The first is real-world data, obtained from various real-world experiments.
    \item The second is synthetic data, artificially generated to mimic real-world patterns. Synthetic data is generated for use instead of real data. Such datasets are of special interest in applications where the amount of data required is much larger than that which is available, or where privacy considerations are crucial and strict, such as in the healthcare domain.
    \item The third type are toy datasets, used for demonstration and visualization purposes. Typically they are artificially generated; often there is no need to represent real-world data patterns.
\end{enumerate}

The foundation of Deep Learning utilization is the availability of data to teach the system about pattern identification.
The effectiveness of the model depends on the quality of the data.
Despite the successful implementation of universal language modeling techniques such as BERT~\cite{sun2019utilizing},~however, such models can be used solely for pre-training the models.
Afterward, the model needs to be trained on the data associated with the desired task.
Henceforth, based on the everyday demands in different machine domains such as NLP, creating new datasets is crucial.

On the other hand, creating new datasets is not usually an easy matter.
Informally speaking, the newly created dataset should be: the right data to train on, sufficient for the evaluation, and accurate to work on.
Answering the questions of ``what is the meaning of right and accurate data'' is highly application-based.
Basically, the data should have sufficient information, which depends on the quality and quantity of the data.

To create a dataset, the first step is always asking ``what are we trying to do and what problem do we need to solve?'' and ``what kind of data do we need and how much of it is required?''
The next step is to create training and testing portions.
The training data set is used to train a model to know how to find the connections between the inputs and the associated outputs.
The test data set is used to assess the intelligence of the machine, i.e., how well the trained model can operate on the unseen test samples.
Next, we must conduct data preparation to make sure the data and its format is simple and understandable for human experts.
After that, the issue of data accessibility and ownership may arise.
Distribution of data may need to have specific authorizations, especially if we are dealing with sensitive or private data.

Given the aforementioned roadmap, creating proper datasets is complicated and of great importance.
That's why few datasets are frequently chosen by the researchers and developers for benchmarking.
A summary of widely used benchmark datasets is provided in Table~\ref{table:datasets}.

\begin{table*}[ht]\caption[Benchmark datasets]{Benchmark datasets.}
\centering
\resizebox{0.95\textwidth}{!}{
\begin{tabular}{|c|c|l|}
\hline
\textbf{Task}                                                                                            & \textbf{Dataset}                                                                                                                         & \textbf{Link}                                                                                                                                                                                                                                                                                                                                                                                                                                                                                                                                                                                                                                                               \\ \hline
Machine Translation                                                                                      & \begin{tabular}[c]{@{}c@{}}WMT 2014 EN-DE\\ WMT 2014 EN-FR\end{tabular}                                                                  & \url{http://www-lium.univ-lemans.fr/~schwenk/cslm\_joint\_paper/}                                                                                                                                                                                                                                                                                                                                                                                                                                                                                                                                                                                     \\ \hline
Text Summarization                                                                                       & \begin{tabular}[c]{@{}c@{}}CNN/DM\\ Newsroom\\ DUC\\ Gigaword\end{tabular}                                                               & \begin{tabular}[c]{@{}l@{}}\url{https://cs.nyu.edu/~kcho/DMQA/}\\ \url{https://summari.es/}\\ \url{https://www-nlpir.nist.gov/projects/duc/data.html}\\ \url{https://catalog.ldc.upenn.edu/LDC2012T21}\end{tabular}                                                                                                                                                                                                                                                                                                                                                                                \\ \hline
\begin{tabular}[c]{@{}c@{}}Reading Comprehension\\ Question Answering\\ Question Generation\end{tabular} & \begin{tabular}[c]{@{}c@{}}ARC\\ CliCR\\ CNN/DM\\ NewsQA\\ RACE\\ SQuAD\\ Story Cloze Test\\ NarativeQA\\ Quasar\\ SearchQA\end{tabular} & \begin{tabular}[c]{@{}l@{}}\url{http://data.allenai.org/arc/}\\ \url{http://aclweb.org/anthology/N18-1140}\\ \url{https://cs.nyu.edu/~kcho/DMQA/}\\ \url{https://datasets.maluuba.com/NewsQA}\\ \url{http://www.qizhexie.com/data/RACE_leaderboard}\\ \url{https://rajpurkar.github.io/SQuAD-explorer/}\\ \url{http://aclweb.org/anthology/W17-0906.pdf}\\ \url{https://github.com/deepmind/narrativeqa}\\ \url{https://github.com/bdhingra/quasar}\\ \url{https://github.com/nyu-dl/SearchQA}\end{tabular} \\ \hline
Semantic Parsing                                                                                         & \begin{tabular}[c]{@{}c@{}}AMR parsing\\ ATIS (SQL Parsing)\\ WikiSQL (SQL Parsing)\end{tabular}                                         & \begin{tabular}[c]{@{}l@{}}\url{https://amr.isi.edu/index.html}\\ \url{https://github.com/jkkummerfeld/text2sql-data/tree/master/data}\\ \url{https://github.com/salesforce/WikiSQL}\end{tabular}                                                                                                                                                                                                                                                                                                                                                                                                                        \\ \hline
Sentiment Analysis                                                                                       & \begin{tabular}[c]{@{}c@{}}IMDB Reviews\\ SST\\ Yelp Reviews\\ Subjectivity Dataset\end{tabular}                         & \begin{tabular}[c]{@{}l@{}}\url{http://ai.stanford.edu/~amaas/data/sentiment/}\\ \url{https://nlp.stanford.edu/sentiment/index.html}\\ \url{https://www.yelp.com/dataset/challenge}\\ \url{http://www.cs.cornell.edu/people/pabo/movie-review-data/}\end{tabular}                                                                                                                                                                                                                                 \\ \hline
Text Classification                                                                                      & \begin{tabular}[c]{@{}c@{}}AG News\\ DBpedia\\ TREC\\ 20 NewsGroup\end{tabular}                                                                         & \begin{tabular}[c]{@{}l@{}}\url{http://www.di.unipi.it/~gulli/AG\_corpus\_of\_news\_articles.html}\\ \url{https://wiki.dbpedia.org/Datasets}\\ \url{https://trec.nist.gov/data.html}\\\url{http://qwone.com/~jason/20Newsgroups/}\end{tabular}                                                                                                                                                                                                                                                                                                                                                                                                                   \\ \hline
Natural Language Inference                                                                               & \begin{tabular}[c]{@{}c@{}}SNLI Corpus\\ MultiNLI\\ SciTail\end{tabular}                                                                 & \begin{tabular}[c]{@{}l@{}}\url{https://nlp.stanford.edu/projects/snli/}\\ \url{https://www.nyu.edu/projects/bowman/multinli/}\\ \url{http://data.allenai.org/scitail/}\end{tabular}                                                                                                                                                                                                                                                                                                                                                                                                                                     \\ \hline
Semantic Role Labeling                                                                                   & \begin{tabular}[c]{@{}c@{}}Proposition Bank\\ OneNotes\end{tabular}                                                                      & \begin{tabular}[c]{@{}l@{}}\url{http://propbank.github.io/}\\ \url{https://catalog.ldc.upenn.edu/LDC2013T19}\end{tabular}                                                                                                                                                                                                                                                                                                                                                                                                                                                                                                                 \\ \hline
\end{tabular}\label{table:datasets}
}
\end{table*}

\section{Deep Learning for NLP Tasks}

\begin{figure*}[ht]
\centering
        \includegraphics[scale=0.5]{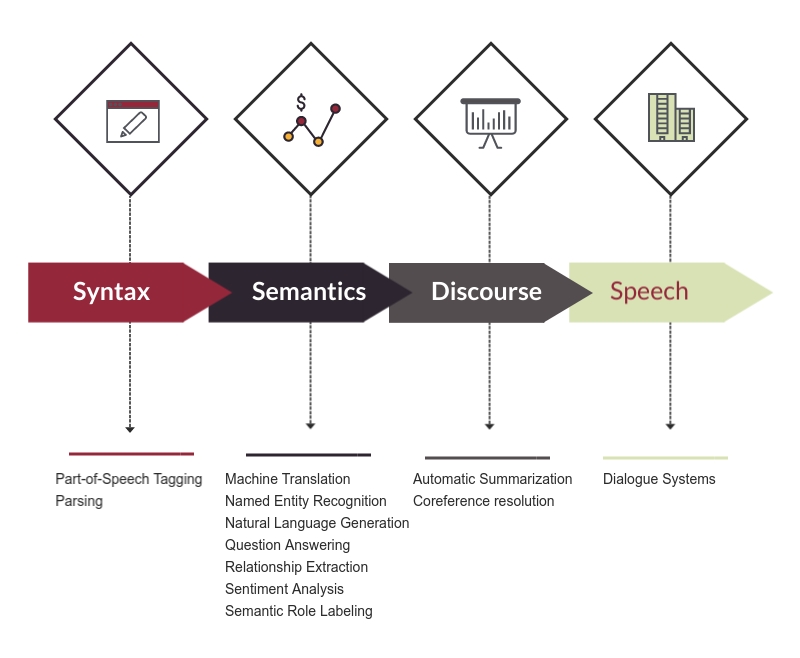}
    \caption{NLP tasks investigated in this study.}
    \label{fig:nlptasks}
\end{figure*}

This section describes NLP applications using deep learning. Fig.~\ref{fig:nlptasks} shows representative NLP tasks~(and the categories they belong to).
A fundamental question is: \textit{"How can we evaluate an NLP algorithm, model, or system?"}
In \cite{resnik2010evaluation}, some of the most common evaluation metrics have been described.
This reference explains the fundamental principles of evaluating NLP systems.

\subsection{Basic Tasks}
\subsubsection{Part-Of-Speech Tagging}

Part-of-Speech tagging is one of the basic tasks in Natural Language Processing.
It is the process of labeling words with their part of speech categories.
Part of speech is leveraged for many crucial tasks such as named entity recognition.
One commonly used dataset for Part-of-Speech tagging is the WSJ corpus\footnote{Penn Treebank Wall Street Journal (WSJ-PTB).}.
This dataset contains over a million tokens and has been utilized widely as a benchmark dataset for the performance assessment of POS tagging systems.
Traditional methods are still performing very well for this task~\cite{manning2011part}. 
However, neural network based methods have been proposed for Part-of-Speech tagging~\cite{kumar2016ask}.

For example, the deep neural network architecture named \textit{CharWNN} has been developed to join word-level and character-level representations using convolutional neural networks for POS tagging~\cite{santos2014learning}.
The emphasis in \cite{santos2014learning} is the importance of character-level feature extraction as their experimental results show the necessity of employing hand-crafted features in the absence of character-level features for achieving the state-of-the-art.
In \cite{huang2015bidirectional}, a wide variety of neural network based models have been proposed for sequence tagging tasks, e.g., LSTM networks, bidirectional LSTM networks, LSTM networks with a CRF\footnote{Conditional Random Field.} layer, etc.
Sequence tagging itself includes part of speech tagging, chunking, and named entity recognition.
Likewise, a globally normalized
transition-based neural network architecture has been proposed for POS-tagging~\cite{andor2016globally}.
State-of-the-art results are summarized in Table ~\ref{table:part-of-speech-tagging}.
In \cite{deshmukh2020deep}, authors propose a bidirectional LSTM to perform parts of speech tagging and show that it performs better than conventional machine learning techniques on the same dataset.
More recently, in \cite{xue2021part}, authors use a pretrained BERT model in combination with one bidirectional LSTM layer and train the latter layer only and outperform the prior state-of-the art POS architectures.

\begin{table}[ht]
\caption[POS tagging state-of-the-art models evaluated on the WSJ-PTB dataset]{POS tagging state-of-the-art models evaluated on the WSJ-PTB dataset.}
\begin{center}
\begin{tabular}{ccccc}
\toprule 
Model & Accuracy \\
\hline
\midrule

\rowcolor{black!0} Character-aware neural language models \cite{liu2018empower} & 97.53\\
\rowcolor{black!5} Transfer Learning + GRU\cite{yang2017transfer} & 97.55\\
\rowcolor{black!10} Bi-directional LSTM + CNNs + CRF\cite{ma2016end} & 97.55\\
\rowcolor{black!15} Adversarial Training + Bi-LSTM \cite{yasunaga2017robust} & 97.59\\
\rowcolor{black!20} Character Composition + Bi-LSTM\cite{ling2015finding} & 97.78\\
\rowcolor{black!25} String Embedding + LSTM\cite{akbik2018contextual} & 97.85\\
\rowcolor{black!30} \textbf{Meta-BiLSTM}  \cite{bohnet2018morphosyntactic} & \textbf{97.96}\\

\bottomrule
\end{tabular}\label{table:part-of-speech-tagging}
\end{center}

\end{table}
\subsubsection{Parsing}

Parsing is assigning a structure to a recognized string.
There are different types of parsing. \textit{Constituency Parsing} refers in particular to assigning a syntactic structure to a sentence.
A greedy parser has been introduced in \cite{legrand2014joint} which performs a syntactic and semantic summary of content using vector representations.
To enhance the results achieved by \cite{legrand2014joint}, the approach proposed in \cite{legrand2016deep} focuses on learning morphological embeddings.
Recently, deep neural network models outperformed traditional algorithms.
State-of-the-art results are summarized in Table~\ref{table:constituency-parsing}.

\begin{table}[ht]
\caption[Constituency parsing state-of-the-art models evaluated on the WSJ-PTB dataset]{Constituency parsing state-of-the-art models evaluated on the WSJ-PTB dataset.}
\begin{center}
\begin{tabular}{ccccc}
\toprule 
Model & Accuracy \\
\hline
\midrule

\rowcolor{black!0} Recurrent neural network grammars~(RNNG)~\cite{kuncoro2016recurrent} & 93.6\\
\rowcolor{black!5} In-order traversal over syntactic trees + LSTM~\cite{liu2017order} & 94.2\\
\rowcolor{black!10} Model Combination and Reranking~\cite{fried2017improving} & 94.6\\
\rowcolor{black!15} \textbf{Self-Attentive Encoder}~\cite{kitaev2018constituency} & \textbf{95.1}\\

\bottomrule
\end{tabular}\label{table:constituency-parsing}
\end{center}

\end{table}

Another type of parsing is called \textit{Dependency Parsing}.
Dependency structure shows the structural relationships between the words in a targeted sentence.~In dependency parsing, phrasal elements and phrase-structure rules do not contribute to the process. Rather, the syntactic structure of the sentence is expressed only in terms of the words in the sentence and the associated relations between the words.

Neural networks have shown their superiority regarding generalizability and reducing the feature computation cost.
In \cite{chen2014fast}, a novel neural network-based approach was proposed for a transition-based dependency parser.
Neural network based models that operate on task-specific transition systems have also been utilized for dependency parsing~\cite{andor2016globally}.
A regularized parser with bi-affine classifiers has been proposed for the prediction of arcs and labels~\cite{dozat2016deep}.
Bidirectional-LSTMs have been used in dependency parsers for feature representation~\cite{kiperwasser2016simple}. A new control structure has been introduced for sequence-to-sequence neural networks based on the stack LSTM and has been used in transition-based parsing~\cite{dyer2015transition}.
\cite{jaf2019deep} presents a transition based multilingual dependency parser which uses a bidirectional LSTM to adapt to target languages. In \cite{zhang2019parsing}, the authors provide a comparison on the state of the art deep learning based parsing methods on a clinical text parsing task.
More recently, in \cite{zhang2020efficient}, a second-order TreeCRF extension was added to the biaffine~\cite{dozat2017deep} parser to demonstrate that structural learning can further improve parsing performance over the state-of-the-art bi-affine models.
\subsubsection{Semantic Role Labeling}

Semantic Role Labeling~(SRL) is the process of identification and classification of text arguments.
It is aimed at the characterization of elements to determine ``who'' did ``what'' to ``whom'' as well as ``how,'' ``where,'' and ``when.''
It identifies the predicate-argument structure of a sentence.
The predicate, in essence, refers to ``what,'' while the arguments consist of the associated participants and properties in the text.
The goal of SRL is to extract the semantic relations between the predicate and the related arguments.

Most of the previously-reported research efforts are based on explicit representations of semantic roles.
Recently, deep learning approaches have achieved the SRL state-of-the-art without taking the explicit syntax representation into consideration~\cite{tan2017deep}.
On the other hand, it is argued that the utilization of syntactic information can be leveraged to improve the performance of syntactic-agnostic\footnote{Note that being syntactic-agnostic does not imply discarding syntactic information. It means they are not explicitly employed.} models~\cite{marcheggiani2017encoding}.
A linguistically-informed self-attention (LISA) model has been proposed to leverage both multi-task learning and self-attention for effective utilization of the syntactic information for SRL~\cite{strubell2018linguistically}.
Current state-of-the-art methods employ joint prediction of predicates and arguments~\cite{he2018jointly}, novel word representation approaches~\cite{peters2018deep}, and self-attention models~\cite{tan2018deep}; see Table~\ref{table:semantic-role-labeling}.

Researchers in \cite{he2019syntax} focus on syntax and contextualized word representation to present a unique multilingual SRL model based on a biaffine scorer, argument pruning and bidirectional LSTMs, (see also \cite{li2019dependency}).

\begin{table}[ht]
\caption[Semantic Role Labeling current state-of-the-art models evaluated on the OntoNotes dataset]{Semantic Role Labeling current state-of-the-art models evaluated on the OntoNotes dataset~\cite{pradhan2013towards}. The accuracy metric is $F_1$ score.}
\begin{center}
\begin{tabular}{ccccc}
\toprule 
Model & Accuracy~($F_1$) \\
\hline
\midrule

\rowcolor{black!0} Self-Attention + RNN~\cite{tan2018deep} & 83.9\\
\rowcolor{black!5} Contextualized Word Representations~\cite{peters2018deep} & 84.6\\
\rowcolor{black!10} \textbf{Argumented Representations + BiLSTM}~\cite{he2018jointly} & \textbf{85.3}\\

\bottomrule
\end{tabular}\label{table:semantic-role-labeling}
\end{center}

\end{table}

\subsection{Text Classification}

The primary objective of text classification is to assign predefined categories to text parts~(which could be a word, sentence, or whole document) for preliminary classification purposes and further organization and analysis.
A simple example is the categorization of given documents as to political or non-political news articles.


The use of CNNs for sentence classification, in which training the model on top of pretrained word-vectors through fine-tuning, has resulted in considerable improvements in learning task-specific vectors~\cite{kim2014convolutional}.
Later, a Dynamic Convolutional Neural Network~(DCNN) architecture -- essentially a CNN with a dynamic k-max pooling method -- was applied to capture the semantic modeling of sentences~\cite{kalchbrenner2014convolutional}.
In addition to CNNs, RNNs have been used for text classification.
An LSTM-RNN architecture has been utilized in \cite{palangi2016deep} for sentence embedding with particular superiority in a defined web search task.
A Hierarchical Attention Network~(HAN) has been utilized to capture the hierarchical structure of text, with a word-level and sentence-level attention mechanism~\cite{yang2016hierarchical}. 

Some models used the combination of both RNNs and CNNs for text classification such as \cite{lai2015recurrent}.
This is a recurrent architecture in addition to max-pooling with an effective word representation method, and demonstrates superiority compared to simple window-based neural network approaches.
Another unified architecture is the C-LSTM proposed in \cite{zhou2015c} for sentence and document modeling in classification.
Current state-of-the-art methods are summarized in Table~\ref{table:text-classification}.
A more recent review of the deep learning based methods for text classification is provided in \cite{minaee2020deep}. The latter focuses on different architectures used for this task, including most recent works in CNN based models, as well as RNN based models, and graph neural networks. In~\cite{zulqarnain2020comparative}, authors provide a comparison between various deep learning methods for text classification, concluding that GRUs and LSTMs can actually perform better than CNN-based models.

\begin{table}[ht]
\caption[The classification accuracy of the state-of-the-art methods, evaluated on the AG News Corpus dataset]{The classification accuracy of state-of-the-art methods, evaluated on the AG News Corpus dataset~\cite{zhang2015character}.}
\begin{center}
\begin{tabular}{ccccc}
\toprule 
Model & Accuracy \\
\hline
\midrule

\rowcolor{black!0} CNN~\cite{conneau2017very} & 91.33\\
\rowcolor{black!5} Deep Pyramid CNN~\cite{johnson2017deep} & 93.13\\
\rowcolor{black!10} CNN~\cite{johnson2016supervised} & 93.43\\
\rowcolor{black!15} \textbf{Universal Language Model Fine-tuning (ULMFiT)}~\cite{howard2018universal} & \textbf{94.99}\\

\bottomrule
\end{tabular}\label{table:text-classification}
\end{center}

\end{table}

\subsection{Information Extraction}

Information extraction identifies structured information from ``unstructured'' data such as social media posts and online news.
Deep learning has been utilized for information extraction regarding subtasks such as \textit{Named Entity Recognition}, \textit{Relation Extraction}, \textit{Coreference Resolution}, and \textit{Event Extraction}.

\subsubsection{Named Entity Recognition}

Named Entity Recognition~(NER) aims to locate and categorize named entities in context into pre-defined categories such as the names of people and places.
The application of deep neural networks in NER has been investigated by the employment of CNN~\cite{collobert2011natural} and RNN architectures~\cite{mesnil2013investigation}, as well as hybrid bidirectional LSTM and CNN architectures~\cite{chiu2015named}.
NeuroNER~\cite{2017neuroner}, a named-entity recognition tool, operates based on artificial neural networks. State-of-the-art models are reported in Table~\ref{table:ner}.
\cite{li2020survey} provides an extensive discussion on recent deep learning methods for named entity recognition. The latter concludes that the work presented in~\cite{baevski2019clozedriven} outperforms other recent models (with an F-score of 93.5 on the CoNLL03 dataset).

\begin{table}[ht]
\caption[State of the art models regarding Name Entity Recognition. Evaluation is performed on the CoNLL-2003 Shared Task dataset]{State of the art models regarding Name Entity Recognition. Evaluation is performed on the CoNLL-2003 Shared Task dataset~\cite{tjong2003introduction}. The evaluation metric is $F_1$ score.}
\begin{center}
\begin{tabular}{ccccc}
\toprule 
Model & Accuracy \\
\hline
\midrule

\rowcolor{black!5} Semi-supervised Sequence Modeling~\cite{clark2018semi} & 92.61\\
\rowcolor{black!10} Google BERT~\cite{devlin2018bert} & 92.8\\
\rowcolor{black!15} \textbf{Contextual String Embeddings}~\cite{akbik2018contextual} & \textbf{93.09}\\

\bottomrule
\end{tabular}\label{table:ner}
\end{center}

\end{table}


\subsubsection{Relation Extraction}

Relation Extraction aims to find the semantic relationships between entity pairs.~The recursive neural network (RNN) model has been proposed for semantic relationship classification by learning compositional vector  representations~\cite{socher2012semantic}.
For relation classification, CNN architectures have been employed as well, by extracting lexical and sentence level features~\cite{zeng2014relation}.
More recently, in~\cite{geng2020semantic}, bidirectional tree-structured LSTMs were shown to perform well for relation extraction. \cite{han2020more} provides a more recent review on relation extraction.





\subsubsection{Coreference Resolution}

Coreference resolution includes identification of the mentions in a context that refer to the same entity.
For instance, the mentions “car,” “Camry,” and “it” could all refer to the same entity. For the first time in \cite{clark2016deep}, Reinforcement Learning~(RL) was applied to coreference resolution.
Current widely used methods leverage an attention mechanism~\cite{lee2018higher}. More recently, in \cite{fei2019end}, authors adopt a reinforcement learning policy gradient approach to coreference resolution and provide state-of-the art performance on the English OntoNotes v5.0 benchmark task. \cite{wu2020corefqa} reformulates coreference resolution as a span prediction task as in question answering and provide superior performance on the CoNLL-2012 benchmark task.



\subsubsection{Event Extraction}

A specific type of extracted information from text is an event. Such extraction may involve recognizing trigger words related to an event and assigning labels to entity mentions that represent event triggers.
Convolutional neural networks have been utilized for event detection; they handle problems with feature-based approaches including exhaustive feature engineering and error propagation phenomena for feature generation~\cite{chen2015event}.
In 2018, Nguyen and Grishman applied graph-CNN (GCCN) where the convolutional operations are applied to syntactically dependent words as well as consecutive words~\cite{nguyen2018graph}; their adding entity information reflected the state-of-the-art using CNN models. 
\cite{zhang2019joint} uses a novel inverse reinforcement learning approach based on generative adversarial networks (imitation learning) to tackle joint entity and event extraction. 
More recently, in \cite{zhao2021novel}, authors proposed a model for document-level event extraction using a combined dependency-based GCN (for local context) and a hypergraph (as an aggregator for global context).


\subsection{Sentiment analysis}

The primary goal in sentiment analysis is the extraction of subjective information from text by contextual mining. 
Sentiment analysis is considered high-level reasoning based on source data.
Sentiment analysis is sometimes called opinion mining, as its primary goal is to analyze human opinion, sentiments, and even emotions regarding products, problems, and varied subjects.
Seminal works on sentiment analysis or opinion mining include \cite{nasukawa2003sentiment,dave2003mining}.
Since 2000, much attention has been given to sentiment analysis, due to its relation to a wide variety of applications~\cite{pathak2020application}, its associations with new research challenges, and the availability of abundant data.
\cite{yadav2020sentiment} provides a more recent review of the sentiment analysis methods relying on deep learning and gives an insightful discussion on the drawbacks as well as merits of deep learning methods for sentiment analysis. 


A critical aspect of research in sentiment analysis is content granularity.
Considering this criterion, sentiment analysis is generally divided into three categories/levels: document level, sentence level, and aspect level. 

\subsubsection{Document-level Sentiment Analysis} 
At the document level, the task is to determine whether the whole document reflects a positive or negative sentiment about exactly one entity.
This differs from opinion mining regarding multiple entries.
The Gated Recurrent Neural Network architecture has been utilized successfully for effectively encoding the sentences' relations in the semantic structure of the document~\cite{tang2015document}.
Domain adaptation has been investigated as well, to deploy the trained model on unseen new sources~\cite{glorot2011domain}.
More recently, in \cite{rao2018lstm} authors provide an LSTM-based model for document-level sentiment analysis that captures semantic relations between sentences. In \cite{rhanoui2019cnn}, authors use a CNN-bidirectional LSTM model to process long texts.

\subsubsection{Sentence-level Sentiment Analysis} 
At the sentence-level, sentiment analysis determines the positivity, negativity, or neutrality regarding an opinion expressed in a sentence.
One general assumption for sentence-level sentiment classification is the existence of only one opinion from a single opinion holder in an expressed sentence.
Recursive autoencoders have been employed for sentence-level sentiment label prediction by learning the vector space representations for phrases~\cite{socher2011semi}.
Long Short-Term Memory (LSTM) recurrent models have also been utilized for tweet sentiment prediction~\cite{wang2015predicting}.
The Sentiment Treebank and Recursive Neural Tensor Networks \cite{socher2013recursive} have shown promise for predicting fine-grained sentiment labels.
\cite{arulmurugan2019classification} provides a cloud-based hybrid machine learning model for sentence level sentiment analysis. More recently in \cite{mevskele2020aldonar}, propose A Lexicalized Domain Ontology and a Regularized Neural Attention model (ALDONAr) for sentence-level aspect-based sentiment analysis that uses a CNN classification module with BERT word embeddings and achieves state-of-the art results.

\subsubsection{Aspect-level Sentiment Analysis} 
Document-level and sentence-level sentiment analysis usually focus on the sentiment itself, not the target of the sentiment, e.g., a product.
Aspect-level sentiment analysis directly targets an opinion, with the assumption of the existence of the sentiment and its target.
A document or sentence may not have a generally positive or negative sentiment, but may have multiple subparts with different targets, each with a positive or negative sentiment.
This can make aspect-level analysis even more challenging than other types of sentiment categorization.

Aspect-level sentiment analysis usually involves \textit{Aspect Sentiment Classification} and \textit{Aspect Extraction}. The former determines opinions on different aspects~(positive, neutral, or negative) while the latter identifies the target aspect for evaluation in context.
As an example consider the following sentence: \textit{``This car is old. It must be repaired and sold!''}. ``This car'' is what is subject to evaluation and must be extracted first. Here, the opinion about this aspect is negative.

For aspect-level sentiment classification, attention-based LSTMs are proposed to connect the aspect and sentence content for sentiment classification~\cite{wang2016attention}.
For aspect extraction, deep learning has successfully been proposed in opinion mining~\cite{ma2018sentic}.~State-of-the-art methods rely on converting aspect-based sentiment analysis to sentence-pair classification tasks~\cite{sun2019utilizing}, post-training approaches~\cite{xu2019bert} on the popular language model BERT~\cite{devlin2018bert}, and employment of pre-trained embeddings~\cite{xu2018double}.
\cite{do2019deep} provides a recent comparative review on aspect-based sentiment analysis. Also recently, \cite{rida2019multi} proposed a dual-attention model which tries to extract the implicit relation between the aspect and opinion terms. In \cite{liang2019novel} authors propose a novel Aspect-Guided Deep Transition model for aspect-based sentiment analysis.

\subsection{Machine Translation}

Machine Translation (MT) is one of the areas of NLP that has been profoundly affected by the advances in deep learning.
The first subsection below explains methods used in the pre-deep learning period, as explained in reference NLP textbooks such as ``Speech and Language Processing"~\cite{jurafsky2008speech}.
The remainder of this section is dedicated to delving into recent innovations in MT which are based on neural networks, started by~\cite{kalchbrenner2013recurrent}. \cite{singh2017machine, yang2020survey} provide reviews on various deep learning architectures used for MT.

\subsubsection{Traditional Machine Translation}

One of the first demonstrations of machine translation happened in 1954~\cite{dostert1955georgetown} in which the authors tried to translate from Russian to English.
This translation system was based on six simple rules, but had a very limited vocabulary.
It was not until the 1990s that successful statistical implementations of machine translation emerged as more bilingual corpora became available~\cite{jurafsky2008speech}. In~\cite{papineni2002bleu} the BLEU score was introduced as a new evaluation metric, allowing more rapid improvement than when the only approach involved using human labor for evaluation.


\subsubsection{Neural Machine Translation}
It was after the success of the neural network in image classification tasks that researchers started to use neural networks in machine translation (NMT).
Around 2013, research groups started to achieve breakthrough results in NMT.
Unlike traditional statistical machine translation, NMT is based on an end-to-end neural network~\cite{bahdanau2014neural}. This implies that there is no need for extensive preprocessing and word alignments. Instead, the focus shifted toward network structure.

\begin{figure}[ht]
\centering
        \includegraphics[totalheight=5cm]{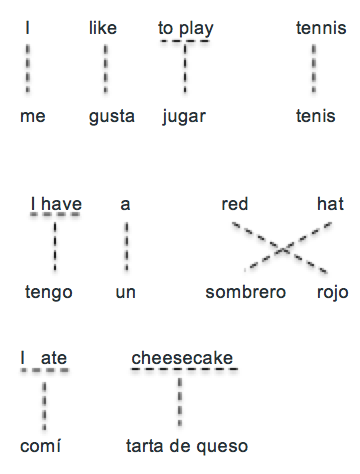}
    \caption{Alignment in Machine Translation}
    \label{fig:align}
\end{figure}

Fig.~\ref{fig:seq-to-seq_with_LSTM} shows an example of an end-to-end recurrent neural network for machine translation.
A sequence of input tokens is fed into the network.
Once it reaches an end-of-sentence (EOS) token, it starts generating the output sequence.
The output sequence is generated in the same recurrent manner as the input sequence until it reaches an end-of-sentence token.
One major advantage of this approach is that there is no need to specify the length of the sequence; the network takes it into account automatically.
In other words, the end-of-sentence token determines the length of the sequence.
Networks implicitly learn that longer input sentences usually lead to longer output sentences with varying length, and that ordering can change.
For instance, the second example in Fig.~\ref{fig:align} shows that adjectives generally come before nouns in English but after nouns in Spanish.
There is no need to explicitly specify this since the network can capture such properties.
Moreover, the amount of memory that is used by NMT is just a fraction of the memory that is used in traditional statistical machine translation~\cite{cho2014properties}. 

~\cite{kalchbrenner2013recurrent} was one of the early works that incorporated recurrent neural networks for machine translation. They were able to achieve a perplexity (a measure where lower values indicate better models) that was 43\% less than the state-of-the-art alignment based translation models.
Their recurrent continuous translation model (RCTM) is able to capture word ordering, syntax, and meaning of the source sentence explicitly.
It maps a source sentence into a probability distribution over sentences in the target language. RCTM estimates the probability $P(f|e)$ of translating a sentence $e = e_1 + ... + e_k$ in the source language to target language sentence $f = f_1 + ... + f_m$.
RCTM estimates $P(f|e)$ by considering source sentence $e$ as well as the preceding words in the target language $f_{1:i-1}$:

\begin{equation}
P(f|e) = \prod_{i=1}^{m} P(f_i|f_{1:i-1},e)
\label{eq:MachineTranslation1}
\end{equation}

\begin{figure}[ht]
\centering
        \includegraphics[totalheight=6cm]{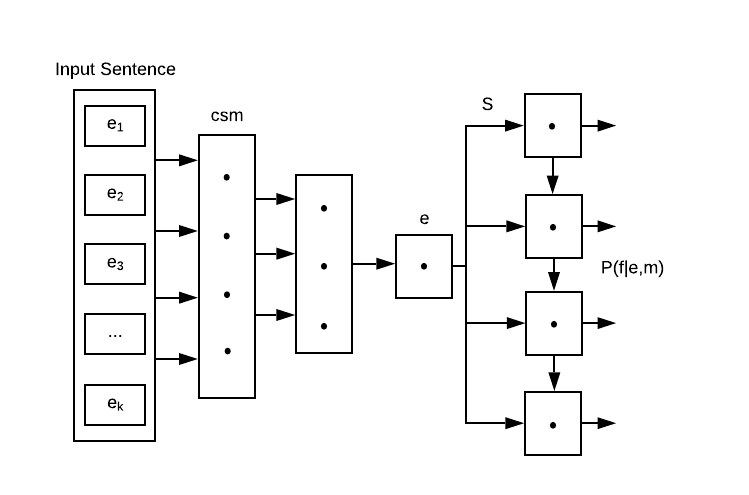}
    \caption{Recurrent Continuous Translation Models (RCTM)~\cite{kalchbrenner2013recurrent}.}
    \label{fig:RCTM}
\end{figure}

The representation generated by RCTM acts on n-grams in the lower layers, and acts more on the whole sentence as one moves to the upper layers.
This hierarchical representation is performed by applying different layers of convolution.
First a continuous representation of each word is generated; i.e., if the sentence is $e = e_1 ... e_k$, the representation of the word $e_i$ will be $v(e_i) \in \mathbb{R}^{q\times1}$.
This will result in sentence matrix $\textbf{E}^e \in \mathbb{R}^{q\times{k}}$ in which $\textbf{E}_{:,i}^e = v(e_i)$.
This matrix representation of the sentence will be fed into a series of convolution layers in order to generate the final representation $\textbf{e}$ for the recurrent neural network. The approach is illustrated in Fig.~\ref{fig:RCTM}.
Equations for the pipeline are as follows. 
\begin{equation}
s = \textbf{S}.csm(e)
\label{eq:MachineTranslation2}
\end{equation}
\begin{equation}
h_1 = \sigma(\textbf{I}.v(f_1) + \textbf{s})
\label{eq:MachineTranslation3}   
\end{equation}
\begin{equation}
h_{i+1} = \sigma(\textbf{R}.h_i + \textbf{I}.v(f_{i+1}) + \textbf{s})
\label{eq:MachineTranslation4}   
\end{equation}
\begin{equation}
o_{i+1} = \textbf{O} . h_i
\label{eq:MachineTranslation5}   
\end{equation}

In order to take into account the sentence length, the authors introduced RCTM II which estimates the length of the target sentence.
RCTM II was able to achieve better perplexity on WMT datasets (see top portion of Table~\ref{table:datasets}) than other existing machine translation systems.

In another line of work, ~\cite{sutskever2014sequence} presented an end-to-end sequence learning approach without heavy assumptions on the structure of the sequence.
Their approach consists of two LSTMs, one for mapping the input to a vector of fixed dimension and another LSTM for decoding the output sequence from the vector.
Their model was able to handle long sentences as well as sentence representations that are sensitive to word order.
As shown in Fig.~\ref{fig:seq-to-seq_with_LSTM}, the model reads "ABC" as an input sequence and produces "WXYZ" as output sequence.
The $<EOS>$ token indicates the end of prediction.
The network was trained by maximizing the log probability of the translation ($\eta$) given the input sequence ($\zeta$).
In other words, the objective function is:
\begin{figure}[ht]
\centering
        \includegraphics[width=\columnwidth]{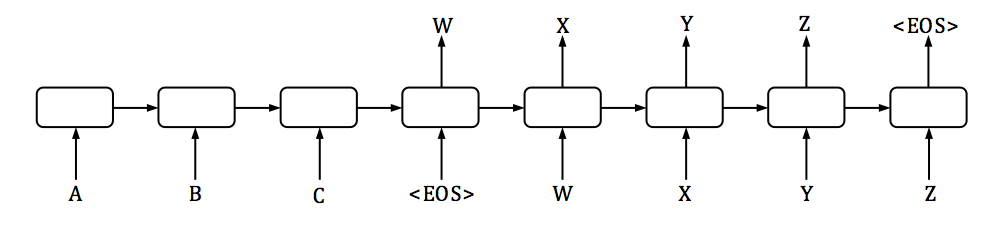}
    \caption{Sequence to sequence learning with LSTM.}
    \label{fig:seq-to-seq_with_LSTM}
\end{figure}

\begin{equation}
1/|\mathcal{D}| \sum_{\substack{(\eta,\zeta) \in \mathcal{D}}} log P(\eta|\zeta)
\label{eq:MachineTranslation6}   
\end{equation}

$\mathcal{D}$ is the training set and $|\mathcal{D}|$ is its size.
One of the novelties of their approach was reversing word order of the source sentence.
This helps the LSTM to learn long term dependencies.

\begin{figure}[ht]
\centering
        \includegraphics[totalheight=6cm]{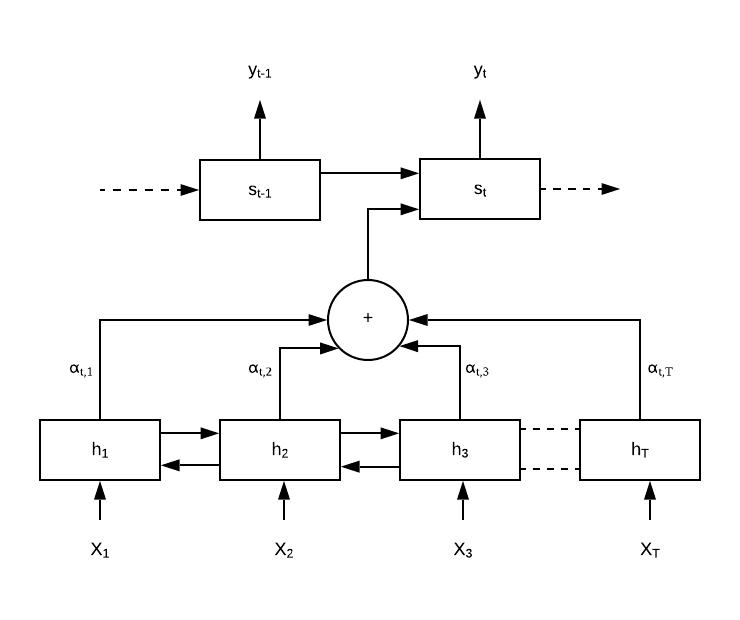}
    \caption{Attention Mechasim for Neural Machine Translation~\cite{bahdanau2014neural}.}
    \label{fig:attention_mechanism_NMT}
\end{figure}

Having a fixed-length vector in the decoder phase is one of the bottlenecks of the encoder-decoder approach. ~\cite{bahdanau2014neural} argues that a network will have a hard time compressing all the information from the input sentence into a fixed-size vector.
They address this by allowing the network to search segments of the source sentence that are useful for predicting the translation.
Instead of representing the input sentence as a fixed-size vector, in \cite{bahdanau2014neural} the input sentence is encoded to a sequence of vectors and a subset of them is chosen by using a method called attention mechanism as shown in Fig.~\ref{fig:attention_mechanism_NMT}.

In their approach $P(y_{i}|y_{1},..., y_{i-1}, X) = g(y_{i-1}, s_{i}, c_{i})$, in which $s_{i} = f(s_{i-1}, y_{i-1}, c_{i})$.
While previously $c$ was the same for all time steps, here $c$ takes a different value, $c_{i}$, at each time step.
This accounts for the attention mechasim (context vector) around that specific time step.
$c_{i}$ is computed according to the following:

$c_{i} = \sum_{j=1}^{T_{x}} \alpha_{ij}h_{j},\ \alpha_{ij} = \frac{exp(e_{ij})}{{\sum_{k=1}^{T_{x}} exp(e_{ik})}},\ e_{ij} = a(s_{i-1}, h_{j})$.

Here $a$ is the alignment model that is represented by a feed forward neural network.
 Also $h_{j} = [\overset{\rightarrow}{h_{j}^T}, \overset{\leftarrow}{h_{j}^T}]$, which is a way to include information both about preceding and following words in $h_{j}$.
 The model was able to outperform the simple encoder-decoder approach regardless of input sentence length.

Improved machine translation models continue to emerge, driven in part by the growth in people's interest and need to understand other languages
Most of them are variants of the end-to-end decoder-encoder approach. For example, ~\cite{wu2016google} tries to deal with the problem of rare words.
Their LSTM network consists of encoder and decoder layers using residual layers along with the attention mechanism.
Their system was able to decrease training time, speed up inference, and handle translation of rare words.
Comparisons between some of the state-of-the-art neural machine translation models are summarized in Table~\ref{table:machine-translation-state-of-the-art}.

\begin{table}[ht]
\caption[The machine translation state-of-the-art models evaluated on the English-German dataset of ACL 2014 Ninth Workshop on Statistical Machine TRranslation]{The machine translation state-of-the-art models evaluated on the \textit{English-German dataset of ACL 2014 Ninth Workshop on Statistical Machine TRranslation}. The evaluation metric is $BLEU$ score.}
\begin{center}
\begin{tabular}{ccccc}
\toprule 
Model & Accuracy \\
\hline
\midrule

\rowcolor{black!0}Convolutional Seq-to-Seq ~\cite{gehring2017convolutional} & 25.2\\
\rowcolor{black!5} Attention Is All You Need~\cite{vaswani2017attention} & 28.4\\
\rowcolor{black!10} Weighted Transformer~\cite{ahmed2017weighted} & 28.9\\
\rowcolor{black!15} Self Attention~\cite{shaw2018self} & 29.2\\
\rowcolor{black!20} DeepL Translation Machine~\footnote{https://www.deepl.com/press.html} & 33.3\\
\rowcolor{black!25} \textbf{Back-translation}~\cite{edunov2018understanding} & \textbf{35.0}\\

\bottomrule
\end{tabular}\label{table:machine-translation-state-of-the-art}
\end{center}

\end{table}

More recently, \cite{aharoni2019massively} provides an interesting single-model implementation of massively multilingual NMT. In \cite{zhu2020incorporating}, authors use BERT to extract contextual embeddings and combine BERT with an attention-based NMT model and provide state-of-the-art results on various benchmark datasets. \cite{liu2020multilingual} proposes mBART which is a seq-to-seq denoising autoencoder and reports that using a pretrained, locked (i.e. no modifications) mBART improves performance in terms of the BLEU point. \cite{cheng2019robust} proposes an interesting adversarial framework for robustifying NMT against noisy inputs and reports performance gains over the Transformer model. \cite{zhang2019bridging} is also an insightful recent work where the authors sample context words from the predicted sequence as well as the ground truth to try to reconcile the training and inference processes. Finally, \cite{yang2020towards} is a successful recent effort to prevent the forgetting that often accompanies in translating pre-trained language models to other NMT task. \cite{yang2020towards} achieves that aim primarily by using a dynamically gated model and asymptotic distillation.

\subsection{Question Answering}

Question answering (QA) is a fine-grained version of Information Retrieval (IR).
In IR a desired set of information has to be retrieved from a set of documents.
The desired information could be a specific document, text, image, etc.
On the other hand, in QA specific answers are sought, typically ones that can be inferred from available documents.
Other areas of NLP such as reading comprehension and dialogue systems intersect with question answering.

Research in computerized question answering has proceeded since the 1960s.
In this section, we present a general overview of question answering system history, and focus on the breakthroughs in the field.
Like all other fields in NLP, question answering was also impacted by the advancement of deep learning ~\cite{bordes2014question}, so we provide an overview of QA in deep learning contexts.
We briefly visit visual question answering as well.

\subsubsection{Rule-based Question Answering}
Baseball ~\cite{green1961baseball} is one of the early works (1961) on QA where an effort was made to answer questions related to baseball games by using a game database.
The baseball system consists of (1) question read-in, (2) dictionary lookup for words in the question, (3) syntactic (POS) analysis of the words in question, (4) content analysis for extracting the input question, and (5) estimating relevance regarding answering the input question.

IBM's ~\cite{ittycheriah2000ibm} statistical question answering system consisted of four major components:
\begin{enumerate}
  \item Question/Answer Type Classification
  \item  Query Expansion/Information Retrieval
  \item  Name Entity Making
  \item  Answer Selection
\end{enumerate}

Some QA systems fail when semantically equivalent relationships are phrased differently. ~\cite{cui2005question} addressed this by proposing fuzzy relation matching based on mutual information and expectation maximization. 

\subsubsection{Question answering in the era of deep learning}
Smartphones (Siri, Ok Google, Alexa, etc.) and virtual personal assistants are common examples of QA systems with which many interact on a daily basis.
While earlier such systems employed rule-based methods, today their core algorithm is based on deep learning.
Table~\ref{tab:QA_Siri} presents some questions and answers provided by Siri on an iPhone.

\begin{table}[ht]
    \centering
    \caption{Typical Question Answering performance based on deep learning.}
    \begin{tabular}{|p{0.45\columnwidth}|p{0.45\columnwidth}|}
    \hline
     \textbf{Question}             & \textbf{Answer} \\ \hline
     Who invented polio vaccine? & \textit{The answer I found is Jonas Salk} \\ \hline
     Who wrote Harry Potter? &  \textit{J.K.Rowling wrote Harry Potter in 1997} \\ \hline
     When was Einstein born?  &  \textit{Albert Einstein was born March 14, 1879} \\ \hline
    \end{tabular}
    \label{tab:QA_Siri}
\end{table}

\begin{figure*}[ht]
    \centering
    \includegraphics[totalheight=6cm]{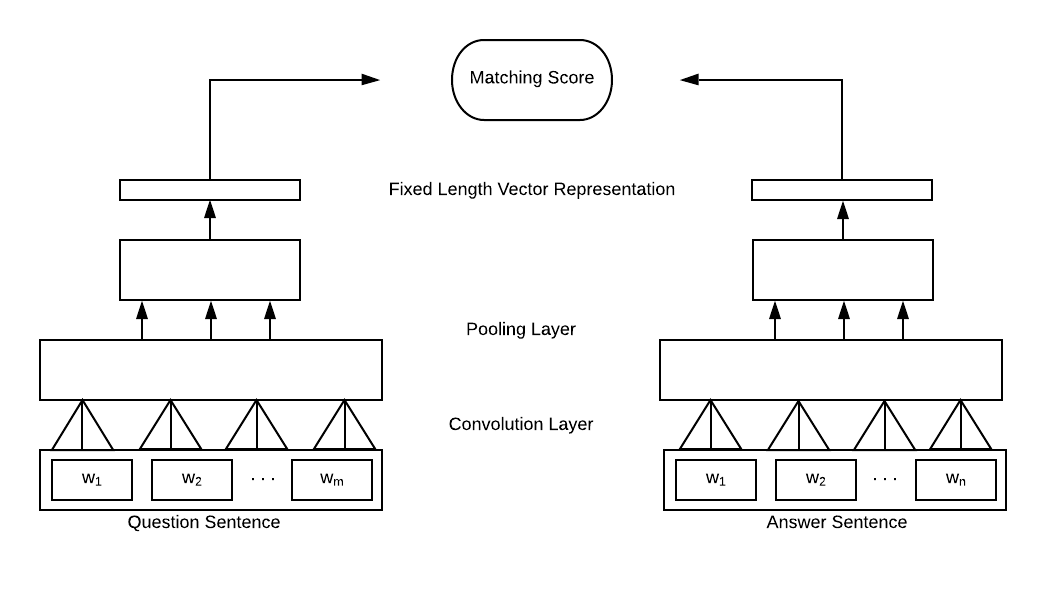}
    \caption{Fixed length vector sentence representation for input Questions and Answers~\cite{qiu2015convolutional}.}
    \label{fig:QASentRep}
\end{figure*}

~\cite{ng2000machine} was one of the first machine learning based papers that reported results on QA for a reading comprehension test.
The system tries to pick a sentence in the database that has an answer to a question, and a feature vector represents each question-sentence pair.
The main contribution of \cite{ng2000machine} is proposing a feature vector representation framework which is aimed to provide information for learning the model.
There are five classifiers (location, date, etc.), one for each type of question.
They were able to achieve accuracy competitive with previous approaches.

As illustrated in Fig. \ref{fig:QASentRep}, \cite{qiu2015convolutional} uses convolutional neural networks in order to encode Question-Answer sentence pairs in the form of fixed length vectors regardless of the length of the input sentence.
Instead of using distance measures like cosine correlation, they incorporate a non-linear tensor layer to match the relevance between question and answer. Equation~\ref{eq:blah_blah1} calculates the matching degree between question $q$ and its corresponding answer $a$. 
\begin{equation}
    s(q,a) = \textbf{u}^T \textbf{f} ( \textbf{v}^T_q \textbf{M}^{[1:r]} \textbf{v}_a + \textbf{V} \begin{bmatrix}
          \textbf{v}_{q} \\
          \textbf{v}_{a}
         \end{bmatrix}  + \textbf{b})
    \label{eq:blah_blah1}
\end{equation}

$\textbf{f}$ is the standard element-wise non-linearity function, $\textbf{M}^{ [1:r] \in R^{n_s \times n_s \times r}}$ is a tensor, $\textbf{V} \in R^{r \times 2n_s}$, $\textbf{b} \in R^r$, $\textbf{u} \in R^r$.

\begin{figure}[ht]
\centering
        \includegraphics[totalheight=4cm]{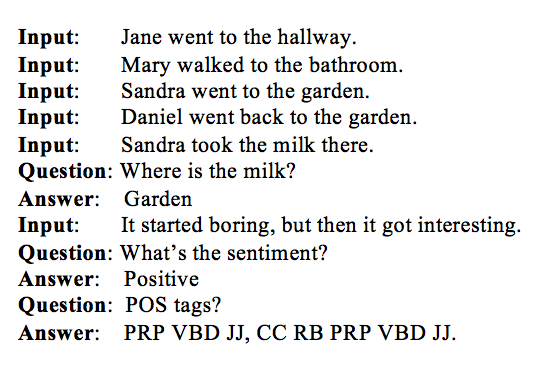}
    \caption{Example of Dynamic Memory Network (DMN) input-question-answer triplet}
\label{fig:DMN}
\end{figure}
The model tries to capture the interaction between question and answer.
Inspired by findings in neuroscience, \cite{kumar2016ask} incorporated episodic memory\footnote{A kind of long-term memory that includes conscious recall of previous activities together with their meaning.} in their Dynamic Memory Network (DMN).~By processing input sequences and questions, DMN forms episodic memories to answer relevant questions.
As illustrated in Fig.~\ref{fig:DMN}, their system is trained based on raw Input-Question-Answer triplets.

\begin{figure}[ht]
\centering
        \includegraphics[totalheight=5cm]{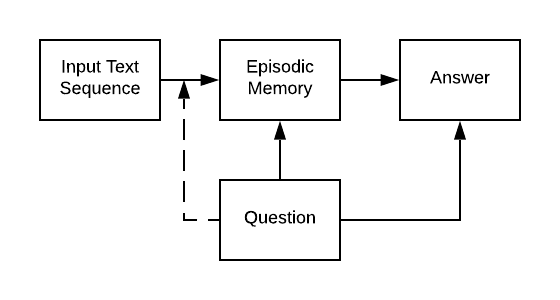}
    \caption{Interaction between four modules of Dynamic Memory Network~\cite{mnih2014recurrent}.}
\label{fig:fourmodule}
\end{figure}

DMN consists of four modules that communicate with each other as shown in Fig.~\ref{fig:fourmodule}. 
The \textbf{input module} encodes raw input text into a distributed vector representation; likewise the \textbf{question  module} encodes a question into its distributed vector representation.
The \textbf{episodic memory module} uses the attention mechanism in order to focus on a specific part of the input module.
Through an iterative process, this module produces a \textit{memory} vector representation that considers the question as well as previous memory.
The \textbf{answer module} uses the final memory vector to generate an answer.
The model improved upon state-of-the-art results on tasks such as the ones shown in Fig.~\ref{fig:DMN}.
DMN is one of the architectures that could potentially be used for a variety of NLP applications such as classification, question answering, and sequence modeling.

\cite{xiong2016dynamic} introduced a Dynamic Coattention Network (DCN) in order to address local maxima corresponding to incorrect answers; it is considered to be one of the best approaches to question answering.


\subsubsection{Visual Question Answering}
\begin{figure}[ht]
\centering
        \includegraphics[totalheight=5cm]{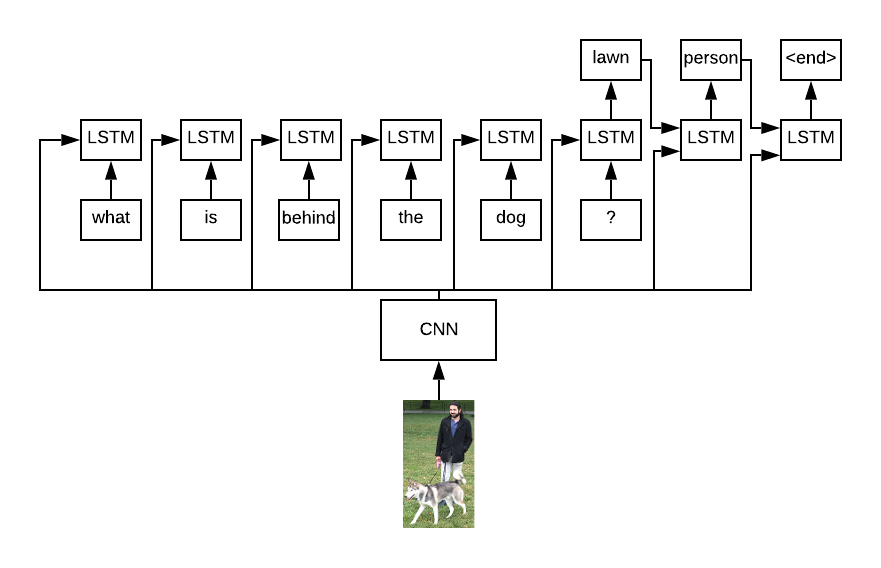}
    \caption{Neural Image Question Answering~\cite{malinowski2015ask}.}
\label{fig:vahid_husky}
\end{figure}

Given an input image, Visual Question Answering (VQA) tries to answer a natural language question about the image \cite{antol2015vqa}.
VQN addresses multiple problems such as object detection, image segmentation, sentiment analysis, etc.
\cite{antol2015vqa} introduced the task of VQA by providing a dataset containing over 250K images, 760K questions, and around 10M answers.
\cite{malinowski2015ask} proposed a neural-based approach to answer the questions regarding the input images.
As illustrated in Fig.~\ref{fig:vahid_husky}, Neural-Image-QA is a deep network consisting of CNN and LSTM.
Since the questions can have multiple answers, the problem is decomposed into predicting a set of answer words $a_{q,x}=\{a_1, a_2, ..., a_{N(q,x)}\}$ from a finite vocabulary set $\nu$ where $N(q,x)$ represents the count of answer words regarding a given question.

\begin{figure}[ht]
\centering
        \includegraphics[totalheight=5cm]{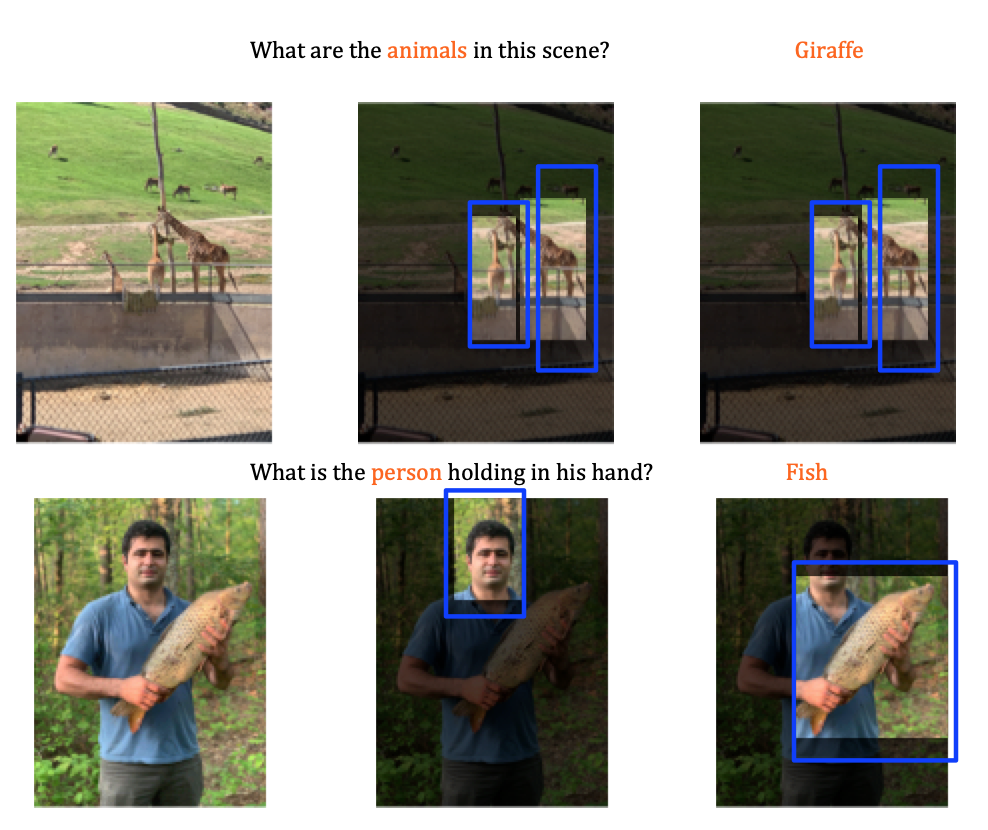}
    \caption{Spatial Memory Network for VQA. Bright Areas are regions the model is attending~\cite{xu2016ask}.}
\label{fig:attentionQA}
\end{figure}
Do humans and computers look at the same regions to answer questions about an image?
\cite{das2017human} tries to answer this question by conducting large-scale studies on \textit{human attention} in VQA.
Their findings show that VQAs do not seem to be looking at the same regions as humans.
Finally, \cite{xu2016ask} incorporates a spatial memory network for VQA.
Fig.~\ref{fig:attentionQA} shows the inference process of their model.
As illustrated in the figure, the specific attention mechanism in their system can highlight areas of interest in the input image. 
\cite{ben2019block} introduces BLOCK, a bilinear fusion model based on superdiagonal tensor decomposition for the VQA task, with state-of-the-art performance and the code made public on github. To improve the generalization of existing models to test data of different distribution, \cite{wu2019selfcritical} introduces a self-critical training objective to help find visual regions of prominent visual/textual correlation with a focus on recognizing influential objects and detecting and devaluing incorrect dominant answers.

\subsection{Document Summarization}
Document summarization refers to a set of problems involving generation of summary sentences given one or multiple documents as input.

Generally, text summarization fits into two categories:
\begin{enumerate}
    \item \textbf{Extractive Summarization}, where the goal is to identify the most salient sentences in the document and return them as the summary.
    \item \textbf{Abstractive Summarization}, where the goal is to generate summary sentences from scratch; they may contain novel words that do not appear in the original document.
\end{enumerate}
Each of these methods has its own advantages and disadvantages.
Extractive summarization is prone to generate long and sometimes overlapping summary sentences; however, the result reflects the author's mode of expression.
Abstractive methods generate a shorter summary but they are hard to train.

There is a vast amount of research on the topic of text summarization using extractive and abstractive methods.
As one of the earliest works on using neural networks for extractive summarization, ~\cite{nallapati2017summarunner} proposed a framework that used a ranking technique to extract the most salient sentences in the input.
This model was improved by~\cite{narayan2018ranking} which used a document-level encoder to represent sentences, and a classifier to rank these sentences.
On the other hand, in abstractive summarization, it was ~\cite{rush2015neural} which, for the first time, used attention over a sequence-to-sequence (seq2seq) model for the problem of headline generation.
However, since simple attention models perform worse than extractive models, therefore more effective attention models such as graph-based attention~\cite{tan2017abstractive} and transformers~\cite{vaswani2017attention} have been proposed for this task.
To further improve abstractive text summarization models, ~\cite{nallapati2016abstractive} proposed the first pointer-generator model and applied it to the DeepMind QA dataset~\cite{hermann2015teaching}.
As a result of this work, the CNN/Daily Mail dataset emerged which is now one of the widely used datasets for the summarization task.
A copy mechanism was also adopted by~\cite{gu2016incorporating} for similar tasks.
But their analysis reveals a key problem with attention-based encoder-decoder models: they often generate unusual summaries consisting of repeated phrases.
Recently, ~\cite{see2017get} reached state-of-the-art results on the abstractive text summarization using a similar framework.
They alleviated the unnatural summaries by avoiding generating unknown tokens and replacing these words with tokens from the input article.
Later, researchers moved their focus to methods that use sentence-embedding to first select the most salient sentence in the document and then change them to make them more abstractive~\cite{chen2018fast,zhou2018neural}.
In these models, salient sentences are extracted first and then a paraphrasing model is used to make them abstractive.
The extraction employs a sentence classifier or ranker while the abstractor tries to remove the extra information in a sentence and present it as a shorter summary.
Fast-RL~\cite{chen2018fast} is the first framework in this family of works.
In Fast-RL, the extractor is pre-trained to select salient sentences and the abstractor is pre-trained using a pointer-generator model to generate paraphrases.
Finally, to merge these two non-differentiable components, they propose using Actor-Critic Q-learning methods in which the actor receives a single document and generates the output while the critic evaluates the output based on comparison with the ground-truth summary.

Though the standard way to evaluate the performance of summarization models is with ROUGE~\cite{lin2004rouge} and BLEU~\cite{papineni2002bleu}, there are major problems with such measures.
For instance, the ROUGE measure focuses on the number of shared n-grams between two sentences.
Such a method incorrectly assigns a low score to an abstractive summary that uses different words yet provides an excellent paraphrase that humans would rate highly.
Clearly, better automated evaluation methods are needed in such cases.

There are additional problems with current summarization models.
Shi \textit{et al.}~\cite{shi2018neural} provides a comprehensive survey on text summarization.

\cite{ma2020multi} provides a recent survey on summarization methods. \cite{abdi2021hybrid} provides an advanced composite deep learning model, based on LSTMs and Restricted Boltzmann Machine, for multi-doc opinion summarization. A very influential recent work, \cite{zhang2019hibert}, introduces {\sc Hibert} ({ HI}erachical { B}idirectional { E}ncoder { R}epresentations from { T}ransformers) as a pre-trained initialization for document summarization and report state-of-the-art performance.

\subsection{Dialogue Systems}

\textit{Dialogue Systems} are quickly becoming a principal instrument in human-computer interaction, due in part to their promising potential and commercial value \cite{merdivan2019dialogue}. One application is automated customer service, supporting both online and bricks-and-mortar businesses.
Customers expect an ever-increasing level of speed, accuracy, and respect while dealing with companies and their services.
Due to the high cost of knowledgeable human resources, companies frequently turn to intelligent conversational machines.
Note that the phrases \textit{conversational machines} and \textit{dialogue machines} are often used interchangeably.

\begin{figure*}[t]
  \centering
  \includegraphics[width=0.7\textwidth,scale=0.5]{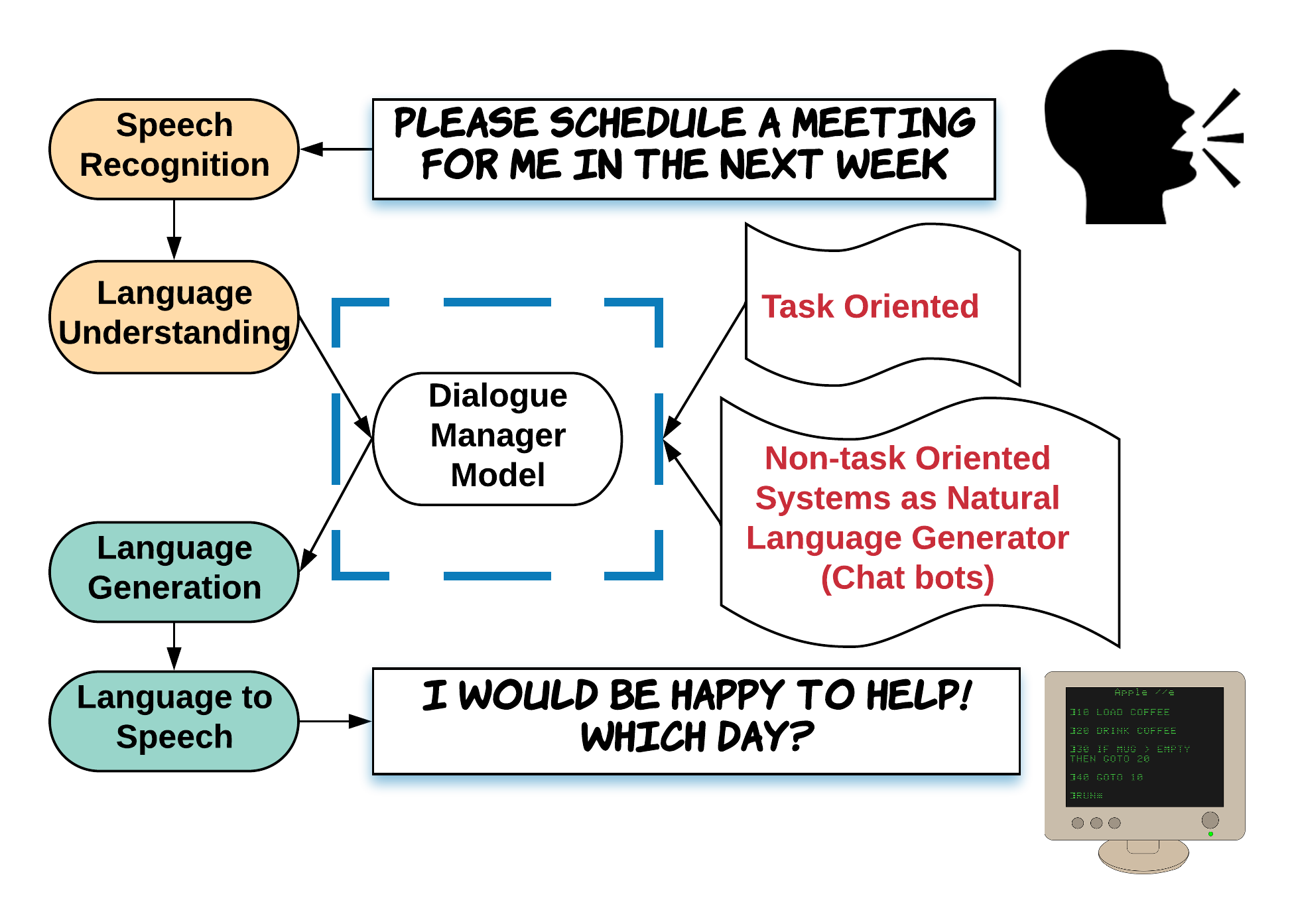}
  \caption{The framework of a dialogue system. A dialogue system can be task oriented or used for natural language generation based on the user input which is also known as a chat bot.}\label{fig:dialoguemachine}
\end{figure*}

Dialogue systems are usually \textit{task-based} or \textit{non-task-based}~(Fig.~\ref{fig:dialoguemachine}).
Though there might be Automatic Speech Recognition~(ASR) and Language-to-Speech~(L2S) components in a dialogue system, the discussion of this section is solely about the linguistic components of dialogue systems; concepts associated with speech technology are ignored.

Despite useful statistical models employed in the backend of dialogue systems~(especially in language understanding modules), most deployed dialogue systems rely on expensive hand-crafted and manual features for operation.~Furthermore, the generalizability of these manually engineered systems to other domains and functionalities is problematic.~Hence, recent attention has focused on deep learning for the enhancement of performance, generalizability, and robustness.
Deep learning facilitates the creation of end-to-end task-oriented dialogue systems, which enriches the framework to generalize conversations beyond annotated task-specific dialogue resources.

\subsubsection{Task-based Systems}

The structure of a task-based dialogue system usually consists of the following elements:

\begin{itemize}
    \item \textit{Natural Language Understanding~(NLU)}: This component deals with understanding and interpreting user's spoken context by assigning a constituent structure to the spoken utterance~(e.g., a sentence) and captures its syntactic representation and semantic interpretation,~to allow the back-end operation/task. NLU is usually leveraged regardless of the dialogue context.
    \item \textit{Dialogue Manager~(DM)}: The generated representation by NLU would be handled by the dialogue manager, which investigates the context and returns a reasonable semantic-related response. 
    \item \textit{Natural Language Generation (NLG)}: The natural language generation (NLG) component produces an utterance based on the response provided by the DM component.
\end{itemize}

The general pipeline is as follows: NLU module (i.e., semantic decoder) transforms the output of the speech recognition module to some dialogue elements.
Then the DM processes these dialogue elements and provides a suitable response which is fed to the NLG for response generation.
The main pipeline in NLU is to classify the user query domain and user intent, and fill a set of slots to create a semantic frame.
It is usually customary to perform the intent prediction and the slot filling simultaneously~\cite{hakkani2016multi}.
Most of the task-oriented dialogue systems employ slot-filling approaches to classify user intent in the specific domain of the conversation.
For this aim, having predefined tasks is required; this depends on manually crafted states with different associated slots.
Henceforth, a designed dialogue system would be of limited or no use for other tasks.

Recent task-oriented dialogue systems have been designed based on deep reinforcement learning, which provided promising results regarding performance~\cite{toxtli2018understanding},~domain adaptation~\cite{ilievski2018goal},~and dialogue generation~\cite{li2016deep}.
This was due to a shift towards end-to-end trainable frameworks to design and deploy task-oriented dialogue systems.
Instead of the traditionally utilized pipeline, an end-to-end framework incorporates and uses a single module that deals with external databases.
Despite the tractability of \textit{end-to-end dialogue systems} (i.e., easy to train and simple to engineer), due to their need for interoperability with external databases via queries, they are not well-suited for task-oriented settings.
Some approaches to this challenge include converting the user input into internal representations~\cite{wen2016network}, combining supervised and reinforced learning~\cite{williams2016end}, and extending the memory network approach~\cite{sukhbaatar2015end} for question-answering to a dialog system~\cite{bordes2016learning}. 

\subsubsection{Non-task-based Systems}

As opposed to task-based dialogue systems, the goal behind designing and deploying non-task-based dialogue systems is to empower a machine with the ability to have a natural conversation with humans~\cite{ritter2011data}.
Typically, chatbots are of one of the following types: \textit{retrieval-based methods} and \textit{generative methods}.
Retrieval-based models have access to information resources and can provide more concise, fluent, and accurate responses.
However, they are limited regarding the variety of responses they can provide due to their dependency on backend data resources.
Generative models, on the other hand, have the advantage of being able to produce suitable responses when such responses are not in the corpus.
However, as opposed to retrieval-based models, they are more prone to grammatical and conceptual mistakes arising from their generative models.

Retrieval-based methods select an appropriate response from the candidate responses.
Therefore, the key element is the query-response operation.
In general, this problem has been formulated as a search problem and uses IR techniques for task completion~\cite{ji2014information}.
Retrieval-based methods usually employ either \textit{Single-turn Response Matching} or \textit{Multi-turn Response Matching}.
In the first type, the current query~(message) is solely used to select a suitable response~\cite{hu2014convolutional}.
The latter type takes the current message and previous utterances as the system input and retrieves a response based on the instant and temporal information.
The model tries to choose a response which considers the whole context to guarantee conversation consistency.
An LSTM-based model has been proposed \cite{lowe2015ubuntu} for context and response vectors creation.
In \cite{yan2016learning}, various features and multiple data inputs have been incorporated to be ingested using a deep learning framework.
Current base models regarding retrieval-based chatbots rely on multi-turn response selection augmented by an attention mechanism and sequence matching~\cite{zhou2018multi}.

\textit{Generative models} don’t assume the availability of pre-defined responses.
New responses are produced from scratch and are based on the trained model.
Generative models are typically based on sequence to sequence models and map an input query to a target element as the response.
In general, designing and implementing a dialogue agent to be able to converse at the human level is very challenging.
The typical approach usually consists of learning and imitating human conversation.
For this goal, the machine is generally trained on large corpora of conversations.
However, this does not directly remedy the issue of \textit{encountering out-of-corpus conversation}.
The question is: \textit{How can an agent be taught to generate proper responses to conversations that it never has seen?}
It must handle content that is not exactly available in the data corpus that the machine has been trained on, due to the lack of content matching between the query and the corresponding response, resulting from the wide range of plausible queries that humans can provide. 

To tackle the aforementioned general problem, some fundamental questions must be answered: \textbf{(1)} What are the core characteristics of a natural conversation? \textbf{(2)} How can these characteristics be measured?  \textbf{(3)} How can we incorporate this knowledge in a machine, i.e., the dialogue system?
Effective integration of these three elements determines the intelligence of a machine.
A qualitative criterion is to observe if the generated utterances can be distinguished from natural human dialogues.
For quantitative evaluation, adversarial evaluation was initially used for quality assessment of sentence generation~\cite{bowman2015generating} and employed for quality evaluation of dialogue systems~\cite{kannan2017adversarial}.
Recent advancements in sequence to sequence modeling encouraged many research efforts regarding natural language generation~\cite{vinyals2015neural}.
Furthermore, deep reinforcement learning yields promising performance in natural language generation~\cite{li2016deep}.

\subsubsection{Final note on dialogue systems}

Despite remarkable advancements in AI and much attention dedicated to dialogue systems, in reality, successful commercial tools, such as Apple's Siri and Amazon's Alexa, still heavily rely on handcrafted features.
It still is very challenging to design and train data-driven dialogue machines given the complexity of the natural language, the difficulties in framework design, and the complex nature of available data sources.

\section{Conclusion}

In this article, we presented a comprehensive survey of the most distinguished works in Natural Language Processing using deep learning.
We provided a categorized context for introducing different NLP core concepts, aspects, and applications, and emphasized the most significant conducted research efforts in each associated category.
Deep learning and NLP are two of the most rapidly developing research topics nowadays.
Due to this rapid progress, it is hoped that soon, new effective models will supersede the current state-of-the-art approaches.
This may cause some of the references provided in the survey to become dated, but those are likely to be cited by new publications that describe improved methods

Neverthless, one of the essential characteristics of this survey is its educational aspect, which provides a precise understanding of the critical elements of this field and explains the most notable research works.
Hopefully, this survey will guide students and researchers with essential resources, both to learn what is necessary to know, and to advance further the integration of NLP with deep learning.

\ifCLASSOPTIONcaptionsoff
  \newpage
\fi

\bibliographystyle{ieeetr}
\bibliography{main}

\end{document}